%% file: main.tex
\pdfoutput=1
\documentclass{article}
\usepackage{iclr2027_conference}
\usepackage{times}
\usepackage{float}
\usepackage{capt-of}
\usepackage[T1]{fontenc}
\usepackage[utf8]{inputenc}
\usepackage{amsmath,amssymb,amsthm,booktabs,graphicx,multirow,tabularx,array}
\usepackage{microtype,xcolor,xurl,hyperref,needspace,placeins,wrapfig}
\usepackage[export]{adjustbox}
\usepackage[most]{tcolorbox}
\usepackage[inline]{enumitem}
\usepackage[normalem]{ulem}

\definecolor{myblue}{HTML}{1F77B4}%
\definecolor{myvermilion}{HTML}{D62728}%
\newtcolorbox{callout}[1][]{%
  enhanced, breakable, parbox=false,
  colframe=myblue, colback=myblue!8, coltitle=white,
  left=5pt, right=5pt, top=3pt, bottom=3pt,
  toprule=2pt, titlerule=1pt, leftrule=1pt, rightrule=1pt, bottomrule=1pt,
  #1}
\newcommand{\takeaway}[2]{\begin{callout}[title={\textbf{Takeaway: #1}}]#2\end{callout}}

\usepackage{tikz}
\usetikzlibrary{arrows.meta,positioning,calc}
\hypersetup{colorlinks=true,linkcolor=myblue,citecolor=myblue,urlcolor=myblue,pdftitle={Success Rates Are Not Model Properties: Controlled Diagnostics for VLA Evaluation},pdfauthor={Kohsuke Ide, Hiroki Nishizawa, Ryousuke Yamada, Yue Qiu, Yoshihiro Fukuhara, Hirokatsu Kataoka, Yutaka Satoh}}
\newtheorem{proposition}{Proposition}
\newcommand{\E}{\mathbb E}

\newcommand{\piz}{\pi_0}
\newcommand{\pif}{\pi_{0.5}}
\newcommand{\Fref}{F_{\mathrm{ref}}}

\input{tables/numbers}

\IfFileExists{tables/ordering_numbers.tex}{\input{tables/ordering_numbers}}{}
\IfFileExists{tables/exemplar_pairs.tex}{

\input{tables/exemplar_pairs}}{}
\IfFileExists{tables/capability_counterexample.tex}{

\input{tables/capability_counterexample}}{}
\IfFileExists{tables/pairedrng_numbers.tex}{

\input{tables/pairedrng_numbers}}{}
\IfFileExists{tables/census_numbers.tex}{\input{tables/census_numbers}}{}
\IfFileExists{tables/state_numbers.tex}{\input{tables/state_numbers}}{}
\IfFileExists{tables/probe_integrity_numbers.tex}{\input{tables/probe_integrity_numbers}}{}
\IfFileExists{tables/demo_distance_numbers.tex}{\input{tables/demo_distance_numbers}}{}
\IfFileExists{tables/normalization_control_numbers.tex}{\input{tables/normalization_control_numbers}}{}
\IfFileExists{tables/goal_instruction_numbers.tex}{

\input{tables/goal_instruction_numbers}}{}
\IfFileExists{tables/goal_summary_numbers.tex}{\input{tables/goal_summary_numbers}}{}
\IfFileExists{tables/field_identification_numbers.tex}{

\input{tables/field_identification_numbers}}{}
\IfFileExists{tables/episode_numbers.tex}{

\input{tables/episode_numbers}}{}
\IfFileExists{tables/roles_numbers.tex}{\input{tables/roles_numbers}}{}
\IfFileExists{tables/reversal_numbers.tex}{

\input{tables/reversal_numbers}}{}
\IfFileExists{tables/horizon_numbers.tex}{

\input{tables/horizon_numbers}}{}
\IfFileExists{tables/state_horizon_numbers.tex}{\input{tables/state_horizon_numbers}}{}
\IfFileExists{tables/horizon_curve_numbers.tex}{

\input{tables/horizon_curve_numbers}}{}
\IfFileExists{tables/length_numbers.tex}{\input{tables/length_numbers}}{}
\IfFileExists{tables/residual_numbers.tex}{\input{tables/residual_numbers}}{}
\IfFileExists{tables/residual_scale_numbers.tex}{

\input{tables/residual_scale_numbers}}{}
\IfFileExists{tables/oft_common_set.tex}{\input{tables/oft_common_set}}{}
\IfFileExists{tables/witness_numbers.tex}{\input{tables/witness_numbers}}{}
\IfFileExists{tables/second_reference.tex}{

\input{tables/second_reference}}{}
\IfFileExists{tables/support_gap_numbers.tex}{\input{tables/support_gap_numbers}}{}
\IfFileExists{tables/suffix_token_numbers.tex}{\input{tables/suffix_token_numbers}}{}
\IfFileExists{tables/suffix_zero_numbers.tex}{

\input{tables/suffix_zero_numbers}}{}
\IfFileExists{tables/instruction_subgroups.tex}{

\input{tables/instruction_subgroups}}{}
\IfFileExists{tables/gap_horizon_numbers.tex}{\input{tables/gap_horizon_numbers}}{}
\IfFileExists{tables/path_numbers.tex}{\input{tables/path_numbers}}{}
\IfFileExists{tables/inversion_numbers.tex}{

\input{tables/inversion_numbers}}{}
\IfFileExists{tables/suite_numbers.tex}{

\input{tables/suite_numbers}}{}
\IfFileExists{tables/suffix_numbers.tex}{

\input{tables/suffix_numbers}}{}
\IfFileExists{tables/suffix_names.tex}{\input{tables/suffix_names}}{}
\IfFileExists{tables/simplerenv_numbers.tex}{

\input{tables/simplerenv_numbers}}{}
\IfFileExists{tables/instr_horizon_numbers.tex}{

\input{tables/instr_horizon_numbers}}{}
\IfFileExists{tables/instruction_interval_numbers.tex}{\input{tables/instruction_interval_numbers}}{}
\IfFileExists{tables/horizon_endpoint_numbers.tex}{

\input{tables/horizon_endpoint_numbers}}{}
\IfFileExists{tables/horizon_short_numbers.tex}{\input{tables/horizon_short_numbers}}{}
\IfFileExists{tables/h5_numbers.tex}{\input{tables/h5_numbers}}{}
\IfFileExists{tables/published_parity_numbers.tex}{\input{tables/published_parity_numbers}}{}
\IfFileExists{tables/real_robot_numbers.tex}{\input{tables/real_robot_numbers}}{}
\IfFileExists{tables/gap_change_numbers.tex}{

\input{tables/gap_change_numbers}}{}
\IfFileExists{tables/intro_numbers.tex}{\input{tables/intro_numbers}}{}
\IfFileExists{tables/v39_numbers.tex}{\input{tables/v39_numbers}}{}
\IfFileExists{tables/app_threshold_numbers.tex}{\input{tables/app_threshold_numbers}}{}
\IfFileExists{tables/app_reference_margins_numbers.tex}{\input{tables/app_reference_margins_numbers}}{}
\IfFileExists{tables/app_states_extra_numbers.tex}{\input{tables/app_states_extra_numbers}}{}
\IfFileExists{tables/app_instruction_extra_numbers.tex}{\input{tables/app_instruction_extra_numbers}}{}
\newif\ifpinnedpending \pinnedpendingfalse
\newcommand{\TBD}{\textcolor{red}{\textbf{TBD}}}
\newcommand{\pendingvalue}[1]{\ifpinnedpending\TBD\else#1\fi}
\newcommand{\pendingsized}[1]{\ifpinnedpending{\sbox0{#1}\setlength{\fboxsep}{0pt}\setlength{\fboxrule}{0.8pt}\fcolorbox{red}{white}{\parbox[c][\dimexpr\ht0+\dp0-1.6pt][c]{\dimexpr\wd0-1.6pt}{\centering\TBD\\[2pt]{\footnotesize fixed-order rerun}}}}\else#1\fi}

\newcommand{\pendinggraphics}[2][]{\pendingsized{\includegraphics[#1]{#2}}}
\title{Success Rates Are Not Model Properties:\\Controlled Diagnostics for VLA Evaluation}
\iclrfinalcopy
\author{\makebox[\dimexpr\textwidth-2\tabcolsep\relax][c]{\begin{tabular}[t]{@{}>{\normalfont}c@{}}
\textbf{Kohsuke Ide}$^{1,2}$ \quad \textbf{Hiroki Nishizawa}$^{1,3}$ \quad \textbf{Ryousuke Yamada}$^{1,4}$ \quad \textbf{Yue Qiu}$^{1}$ \\
\textbf{Yoshihiro Fukuhara}$^{1,5}$ \quad \textbf{Hirokatsu Kataoka}$^{1,6}$ \quad \textbf{Yutaka Satoh}$^{1,2}$ \\[0.25em]
$^{1}$National Institute of Advanced Industrial Science and Technology (AIST) \\
$^{2}$University of Tsukuba \quad $^{3}$Waseda University \\
$^{4}$Fundamental AI Lab, University of Technology Nuremberg \quad $^{5}$CADDi \\
$^{6}$Visual Geometry Group (VGG), University of Oxford
\end{tabular}}}
\begin{document}
\maketitle
\vspace{-0.15in}
\lhead{Preprint}
\input{sections/abstract}
\input{sections/introduction}
\input{sections/questions}
\input{sections/setup}
\input{sections/results}
\input{sections/related}
\input{sections/discussion}
\label{endofmain}
\clearpage

\section*{Acknowledgments}
This work was conducted by AIST researchers under FRONTia, a Japanese national program led by the Ministry of Economy, Trade and Industry of Japan (METI) and the New Energy and Industrial Technology Development Organization (NEDO), aimed at developing a domestic multimodal foundation model for AI robots and physical AI. This work was supported by the AIST policy-based budget project ``R\&D on Generative AI Foundation Models for the Physical Domain''.

This work was supported by Japan Science and Technology Agency (JST) as part of Adopting Sustainable Partnerships for Innovative Research Ecosystem (ASPIRE), Grant Number JPMJAP2518. We used ABCI 3.0 provided by AIST and AIST Solutions with support from ``ABCI 3.0 Development Acceleration Use''.

This paper is based on results obtained from the BRIDGE Program (R7-H05), implemented by the Cabinet Office, Government of Japan.
\par\vspace{\baselineskip}

\bibliography{references}
\bibliographystyle{iclr2027_conference}
\appendix
\input{appendix/a_details}
\input{appendix/b_states}
\input{appendix/c_instruction}
\input{appendix/d_execution}
\input{appendix/e_comparisons}
\input{appendix/f_toy}
\end{document}

%% file: tables/numbers.tex
\newcommand{\BlindN}{7}
\newcommand{\BlindWideN}{15}
\newcommand{\LossWorstUpper}{$-13.5$}
\newcommand{\AdaptCritUpper}{2.62}
\newcommand{\RefKept}{27}

\newcommand{\InstrOtherMax}{8.2}
\newcommand{\InstrSmolMin}{17.7}
\newcommand{\InstrSmolMax}{40.8}
\newcommand{\InstrSmolPlusMin}{28.4}
\newcommand{\InstrSmolPlusMax}{51.7}
\newcommand{\ObjectSmolPlusBefore}{7.3}
\newcommand{\ObjectSmolPlusAfter}{59.0}

%% file: tables/ordering_numbers.tex
\newcommand{\OrdRevObject}{4}

\newcommand{\OrdRevSpatial}{0}

%% file: tables/capability_counterexample.tex
\newcommand{\CexContrasts}{21}

%% file: tables/pairedrng_numbers.tex
\newcommand{\PRCriterionTested}{9}

%% file: tables/census_numbers.tex
\newcommand{\CensusMeta}{8,493}
\newcommand{\CensusPct}{84.7}

\newcommand{\GoalSeverityPop}{7}

%% file: tables/state_numbers.tex
\newcommand{\StatePinned}{6}
\newcommand{\StateTasks}{10}

\newcommand{\StateMaxTargetRangeCM}{1.9}

%% file: tables/probe_integrity_numbers.tex
\newcommand{\ProbeUncleanTasks}{1, 2 and 5}

\newcommand{\ProbeCleanPairs}{50}
\newcommand{\ProbeCleanLower}{50}

\newcommand{\ProbeCleanMeanShift}{1.1}

%% file: tables/demo_distance_numbers.tex
\newcommand{\DemoHausdorffMM}{4.8}
\newcommand{\FixedHausdorffMM}{0.7}
\newcommand{\ProbeDemoMinCM}{2.1}
\newcommand{\ProbeDemoMaxCM}{12.0}

%% file: tables/normalization_control_numbers.tex
\newcommand{\NcLossMin}{46.8}
\newcommand{\NcLossMax}{92.2}
\newcommand{\NcVsOrigMax}{5.3}
\newcommand{\NcTaskEightB}{74.3}
\newcommand{\NcTaskZeroB}{100.0}

\newcommand{\NcZeroMin}{17.3}
\newcommand{\NcZeroMax}{47.5}
\newcommand{\ExTask}{5}

\newcommand{\ExBenchTo}{100.0}
\newcommand{\ExFieldFrom}{93.3}
\newcommand{\ExFieldTo}{34.7}

%% file: tables/goal_instruction_numbers.tex
\newcommand{\GoalSmolVLAPlusSfxTwoDelta}{-51.2}

\newcommand{\GoalPiZeroSfxEightDelta}{-12.4}

\newcommand{\GoalSmolVLASfxTwoDelta}{-61.6}

%% file: tables/goal_summary_numbers.tex
\newcommand{\GoalInstructedMin}{75.8}
\newcommand{\GoalInstructedMax}{98.8}

%% file: tables/roles_numbers.tex
\newcommand{\NumFamilies}{8}
\newcommand{\NumPolicies}{10}
\newcommand{\DefaultHorizonMin}{7}
\newcommand{\DefaultHorizonMax}{50}

%% file: tables/state_horizon_numbers.tex
\newcommand{\HTenXvlaB}{97.8}
\newcommand{\HTenXvlaF}{22.1}
\newcommand{\HTenPiFiveB}{98.2}
\newcommand{\HTenPiFiveF}{89.5}

\newcommand{\HTenFieldGapPts}{67.3}

\newcommand{\HTenLower}{49}

%% file: tables/horizon_curve_numbers.tex
\newcommand{\RouteFiveMin}{7.2}
\newcommand{\RouteFiveMax}{14.4}

\newcommand{\RouteZeroMin}{0.6}
\newcommand{\RouteZeroMax}{12.8}

\newcommand{\VanDeltaMax}{+23.8}

\newcommand{\VanUpMin}{+3.4}

\newcommand{\VanXvlaMin}{-4.4}
\newcommand{\VanXvlaMax}{+2.2}

%% file: tables/length_numbers.tex
\newcommand{\LenMainSuite}{Object}

\newcommand{\LenMin}{11.7}
\newcommand{\LenMax}{51.0}

\newcommand{\ContentMax}{3.7}
\newcommand{\ContentResolved}{2}
\newcommand{\LenCells}{8}
\newcommand{\InstrDistinct}{10{,}002}
\newcommand{\TokLimit}{48}

\newcommand{\TokOver}{one}

\newcommand{\TokFewer}{8{,}493}

%% file: tables/residual_numbers.tex
\newcommand{\ResInstrNoSmolMax}{4.1}
\newcommand{\ResInstrMax}{25.7}

\newcommand{\ResCoreSupportTen}{33.5}

\newcommand{\ResCoreHorizon}{12.4}

%% file: tables/oft_common_set.tex
\newcommand{\OftPairMin}{2,396}
\newcommand{\OftPairMax}{2,590}
\newcommand{\OftSuiteMin}{2,402}
\newcommand{\OftSuiteMax}{2,591}
\newcommand{\OftSubCoverMin}{94.7}
\newcommand{\OftSubCoverMax}{99.8}
\newcommand{\OftSubShift}{0.4}

%% file: tables/support_gap_numbers.tex
\newcommand{\SupRankedRows}{50}

%% file: tables/suffix_token_numbers.tex
\newcommand{\SufTokMin}{42}
\newcommand{\SufTokMax}{44}

%% file: tables/gap_horizon_numbers.tex
\newcommand{\GapPanel}{5}
\newcommand{\GapFieldUp}{4}

\newcommand{\GapPiZeroChangePP}{12.5}

\newcommand{\GapPiFiveChangePP}{36.8}

%% file: tables/suffix_numbers.tex
\newcommand{\SufWorstK}{32}

\newcommand{\SufWorstTo}{0.0}

\newcommand{\SufSmolTo}{0.4}
\newcommand{\SufWorstTwoLoss}{76.0}
\newcommand{\SufSmolTwoLoss}{31.8}
\newcommand{\SufPiFiveLongLoss}{16.4}

\newcommand{\SufSmallMax}{5.8}

%% file: tables/simplerenv_numbers.tex
\newcommand{\SEeps}{24}

\newcommand{\SEbigLo}{5}
\newcommand{\SEbigHi}{30}

\newcommand{\SEbigLoPct}{0}
\newcommand{\SEbigHiPct}{94}

\newcommand{\SECogLo}{1}
\newcommand{\SECogHi}{10}
\newcommand{\SECogLoPct}{8}
\newcommand{\SECogHiPct}{46}

%% file: tables/instr_horizon_numbers.tex
\newcommand{\IHPiZeroSmolPlusFiftyOrig}{+36.5}

\newcommand{\IHPiZeroSmolPlusFiftyRep}{$-$12.4}

%% file: tables/instruction_interval_numbers.tex
\newcommand{\InstrCIDraws}{10,000}

\newcommand{\InstrCIOtherCells}{32}
\newcommand{\InstrCIOtherExclInstance}{14}

%% file: tables/horizon_endpoint_numbers.tex
\newcommand{\HEndCrossAll}{4}
\newcommand{\HEndSupAll}{0}

\newcommand{\HEndXvlaCrossAll}{12}
\newcommand{\HEndXvlaSupAll}{2}

%% file: tables/horizon_short_numbers.tex
\newcommand{\NShortPairs}{20}

%% file: tables/h5_numbers.tex
\newcommand{\HFivePiFiveB}{97.8}
\newcommand{\HFivePiFiveF}{85.5}

%% file: tables/published_parity_numbers.tex
\newcommand{\ParityLoopObject}{99.6}
\newcommand{\ParityDocObject}{99.0}

\newcommand{\ParityLoopSpatial}{96.6}
\newcommand{\ParityDocSpatial}{97.0}

\newcommand{\ParityLoopGoal}{97.8}
\newcommand{\ParityDocGoal}{98.0}

\newcommand{\ParityLoopLong}{95.8}
\newcommand{\ParityDocLong}{96.0}

%% file: tables/real_robot_numbers.tex
\newcommand{\RealTauStd}{0.60}

\newcommand{\RealMMRVStd}{0.064}
\newcommand{\RealMMRVRef}{0.056}
\newcommand{\RealMMRVDiffLo}{$-0.09$}
\newcommand{\RealMMRVDiffHi}{$+0.10$}

%% file: tables/v39_numbers.tex
\newcommand{\ResHorizonAllMin}{7.2}
\newcommand{\ResHorizonAllMax}{12.4}
\newcommand{\ResHorizonNoSmolMin}{5.4}
\newcommand{\ResHorizonNoSmolMax}{8.6}
\newcommand{\NcLossTenMin}{60.2}
\newcommand{\NcLossTenMax}{92.8}

\newcommand{\GoalOtherEightMax}{7.2}

%% file: tables/app_threshold_numbers.tex
\newcommand{\TauNinetyDropped}{3 and 4}
\newcommand{\RefSwapBelowDef}{108}
\newcommand{\RefSwapRowsDef}{120}
\newcommand{\RefSwapBelowTen}{266}
\newcommand{\RefSwapRowsTen}{284}

%% file: tables/app_reference_margins_numbers.tex
\newcommand{\RefMarginCellsTen}{92}
\newcommand{\RefMarginSameTen}{90}

\newcommand{\RefMarginSameSigTen}{84}

\newcommand{\RefMarginCellsDef}{68}
\newcommand{\RefMarginSameDef}{60}

\newcommand{\SelPropPiXvlaLo}{63.8}
\newcommand{\SelPropPiXvlaHi}{73.3}

\newcommand{\SelPropDraws}{10{,}000}
\newcommand{\RefPiXvlaMinThree}{59.5}
\newcommand{\RefPiXvlaMaxThree}{71.0}

%% file: tables/app_states_extra_numbers.tex
\newcommand{\FreshKept}{28}

\newcommand{\FreshKeptOrig}{27}
\newcommand{\FreshRateCorr}{0.99}

\newcommand{\FreshPiXvla}{67.5}
\newcommand{\FreshPiXvlaLo}{63.4}
\newcommand{\FreshPiXvlaHi}{72.1}
\newcommand{\HeldKept}{45}
\newcommand{\HeldCands}{50}

\newcommand{\HeldResolved}{10}
\newcommand{\HeldStdResolved}{5}
\newcommand{\HeldPiXvla}{67.6}
\newcommand{\HeldPiXvlaLo}{64.8}
\newcommand{\HeldPiXvlaHi}{70.4}

\newcommand{\HumPiStd}{-0.9}
\newcommand{\HumPiStdLo}{-2.3}
\newcommand{\HumPiStdHi}{0.4}
\newcommand{\HumPiSel}{6.6}
\newcommand{\HumPiSelLo}{3.5}
\newcommand{\HumPiSelHi}{10.2}
\newcommand{\HumPiSelTen}{4.1}
\newcommand{\HumPiSelTenLo}{1.2}
\newcommand{\HumPiSelTenHi}{6.8}
\newcommand{\HumPiStdWithinFive}{10}
\newcommand{\HumSmolStd}{2.8}
\newcommand{\HumSmolStdLo}{0.2}
\newcommand{\HumSmolStdHi}{5.3}
\newcommand{\HumSmolSel}{-0.9}
\newcommand{\HumSmolSelLo}{-4.4}
\newcommand{\HumSmolSelHi}{2.2}
\newcommand{\HumPiSelTaskA}{7.1}
\newcommand{\HumPiSelTaskF}{6.4}
\newcommand{\HumPiSelTaskI}{14.9}
\newcommand{\SmolScrSelTaskA}{19.7}
\newcommand{\SmolBaseSelTaskA}{19.7}

\newcommand{\SmolScrSelTaskF}{22.5}
\newcommand{\SmolBaseSelTaskF}{22.5}

\newcommand{\SmolScrSelTaskI}{43.3}
\newcommand{\SmolBaseSelTaskI}{46.3}

\newcommand{\SmolScrStdTaskA}{74.2}
\newcommand{\SmolBaseStdTaskA}{80.5}

%% file: tables/app_instruction_extra_numbers.tex
\newcommand{\RWSmolPooled}{36.3}
\newcommand{\RWSmolPlusPooled}{69.1}
\newcommand{\RWSmolFiller}{31.8}
\newcommand{\RWSmolPlusFiller}{76.0}
\newcommand{\RWOthersMax}{4.5}
\newcommand{\PRSmolAll}{21.4}
\newcommand{\PRSmolPlusAll}{36.4}
\newcommand{\PROthersSuiteMax}{8.7}
\newcommand{\PRSeedSDMax}{2.3}

%% file: sections/abstract.tex
\begin{abstract}
Success rates on standard manipulation benchmarks are the primary evidence for comparing
vision-language-action (VLA) models. But a success rate is not a model property: it is measured
under conditions fixed by the benchmark or by the configuration distributed with the model. We show
that three of these conditions, changed with the model fixed, move success by amounts that are large
for some models and small for others. The three are (1)~which initial states are tested, (2)~what
instruction the model receives, and (3)~how many predicted actions it executes between observations (the execution horizon). We evaluate multiple VLA models on LIBERO and LIBERO-Plus, and repeat the execution study in SimplerEnv. (1)~For initial states, X-VLA and $\pi_{0.5}$ both succeed on about 98\% of
LIBERO-Object's standard initial states at the same execution horizon, but on \HTenXvlaF{}\% and
\HTenPiFiveF{}\% of placements outside these states on which a separate VLA model succeeds at a high
rate. After task-specific fine-tuning of $\pi_{0.5}$, success on the same placements falls by more
than the standard score changes, on average for every recipe. (2)~For the instruction, removing the parameters that LIBERO-Plus appends to
most instructions raises the success of the two SmolVLA models by up to \InstrSmolPlusMax{}
percentage points and changes that of the other eight models by at most \InstrOtherMax{} points. For these two models, the delivered text, not just the scene,
drives success. Appending two ordinary words to
standard instructions also lowers their success, even though both models perform another task when
given that task's instruction. (3)~For execution, a horizon of ten instead of each model's default
raises the standard score of five of the six models on LIBERO and also changes success on the
placements from~(1), while in SimplerEnv each tested configuration performs worst at its shortest horizon. Because models respond differently, these conditions move the margins
between models and sometimes reverse which model scores higher, so a comparison made under one
setting does not necessarily hold under another. We give an evaluation procedure that reports each
model's response to these conditions and states when a comparison holds, asking not only how often a
model succeeds, but under which conditions it does.
\end{abstract}

%% file: sections/introduction.tex
\section{Introduction}
Success rates on standard manipulation benchmarks are the primary evidence for progress in
vision-language-action (VLA) models~\citep{liu2023libero,black2024pi0,physicalintelligence2025pi05,kim2025oft}.
A high standard score, the success rate on the benchmark's own initial states and
instructions, establishes performance on the evaluated episodes. However, it does not show
whether the model succeeds from other initial states, tolerates text that does not change the task, or
maintains its success under a different execution horizon. VLA models predict several actions at each
observation, and the execution horizon $h$ is the number of these actions executed before the next observation. A controlled diagnosis changes the initial states, the text or the horizon one at a time for the same model. This paper asks one question:
\begin{center}
\emph{How do evaluation choices change the conclusions drawn from VLA success rates?}
\end{center}

The LIBERO-Object suite illustrates the question. In \StatePinned{} of its \StateTasks{} tasks, the target object never moves across the 50 standard initial states. We translate this object to placements absent
from the standard initial states, retaining only those where a separate VLA model, used as a
reference, succeeds at a high rate (termed reference-selected conditions). At the same execution horizon, X-VLA and $\pif$ both succeed on about 98\% of the standard initial states, but on
\HTenXvlaF{}\% and \HTenPiFiveF{}\% of the reference-selected conditions, respectively. Two models with nearly equal
standard scores thus differ by \HTenFieldGapPts{} points at other placements of the same tasks.

\begin{figure}[t]
\centering
\pendinggraphics[width=\linewidth]{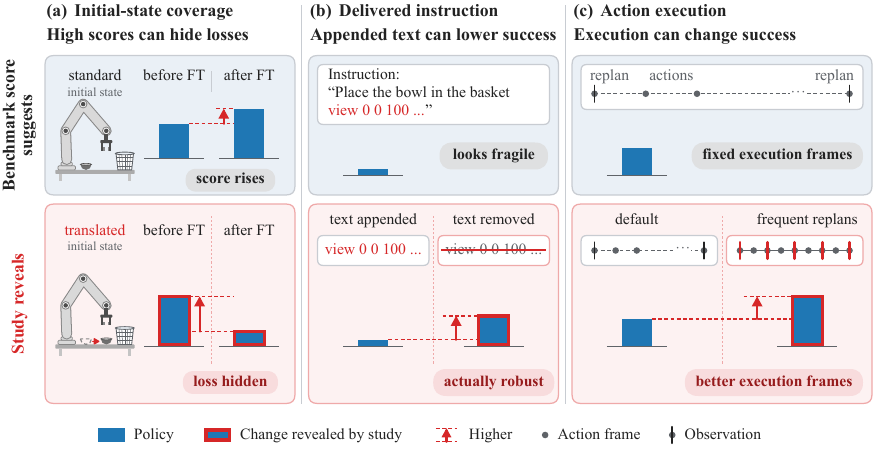}
\caption{\textbf{Three evaluation choices, and what a controlled change of each reveals.} Schematic: (a) compares a model before and after fine-tuning; (b) and (c) change one condition of a fixed model.}
\label{fig:divergence}
\end{figure}

We organize our diagnosis around three conditions fixed by the benchmark or the configuration
distributed with the model, which can be modified at evaluation time without retraining
(Figure~\ref{fig:divergence}). (1)~\emph{Initial-state coverage} asks whether a high standard score
extends to other initial states. (2)~The \emph{delivered instruction}, the text the model actually
receives, asks whether text that does not
change the task alters success. Instructions assembled by software or written by people can carry such
text. (3)~\emph{Action execution} asks how the execution horizon changes the standard score and success
elsewhere. We isolate each factor as a controlled intervention on a fixed model, and refer to the three diagnostics
as the state, instruction and execution studies. For the state study, we also compare a model before
and after task-specific fine-tuning~\citep[e.g.,][]{kim2025oft} on the same initial states, to evaluate how
fine-tuning alters success at other placements.

Applied to multiple VLA models on LIBERO~\citep{liu2023libero}, its
perturbed extension LIBERO-Plus~\citep{fei2025liberoplus} (Table~\ref{tab:roles}), and SimplerEnv~\citep{li2024simpler}, the diagnostics reveal that each condition changes success by model-dependent
amounts, and each exposes something the standard score does not show. (1)~For initial-state coverage,
the standard score measures success from start positions that the training demonstrations also use.
Beyond them, models with similar standard scores differ widely, and fine-tuning can lower success at other placements while the standard score stays high. (2)~For the delivered instruction, a model that follows the
instruction can still fail when task-irrelevant words are added. A robustness benchmark that alters the text alongside
the scene therefore partly measures sensitivity to that text. (3)~For action execution, the
execution horizon changes both the standard score and the gap to other placements, and the favorable
horizon differs across models and simulators. Because responses differ across models, these conditions
move the margins between them and sometimes reverse which model scores higher. Which model is better
therefore depends on the conditions, whereas a benchmark fixes conditions that vary in deployment.

Practitioners know that these conditions matter. VLA models can collapse when object positions change~\citep{zhou2025liberopro}, and shortcuts and overfitting limit what high
scores establish~\citep{jiang2026benchmarking}. Users have reported that LIBERO-Plus appends text to its instructions and measured its cost for one model\footnote{\label{fn:issues}\url{https://github.com/sylvestf/LIBERO-plus/issues}, issues 48, 64 and 65.}.
The number of executed actions is set per model: the public configuration of $\pif$ executes fifty
actions, while an issue report finds that executing ten recovers much of its LIBERO
performance\footnote{\label{fn:lerobot}\url{https://github.com/huggingface/lerobot/issues/2114}}. Knowing that these conditions matter does not show how large each effect is when the model is held fixed and the altered condition is
one on which another model succeeds, whether the effect is the same across models, and therefore whether
a comparison made at one setting holds at another. We measure these quantities for three conditions that
the benchmark or the model's configuration fixes. Our position is methodological: a success rate is a
property of a model together with these conditions, not of the model alone, and it should be reported
with them. Our contributions are as follows:
\begin{itemize}[leftmargin=*,topsep=2pt,itemsep=1pt]
\item \textbf{Controlled measurements of three evaluation conditions.} For fixed models, we measure how
initial-state coverage, the delivered instruction and the execution horizon change success, including
before and after fine-tuning. The effects are large and model-dependent: nearly equal standard scores
hide a \HTenFieldGapPts{}-point difference at other placements, fine-tuning can cut success there from
\ExFieldFrom{}\% to \ExFieldTo{}\% while the standard score stays at \ExBenchTo{}\%, and two
task-irrelevant words lower one model's success by \SufWorstTwoLoss{} points (Section~\ref{sec:findings}).
\item \textbf{An evaluation procedure for comparisons that hold.} We evaluate additional initial states
selected by another model's success and check this choice by repeating it with each other model as the reference,
pair conditions on the same initial states with intervals, and report each model's response to a
condition instead of one rate. This separates comparisons that survive a change of condition from those
that do not (Sections~\ref{sec:comparison} and~\ref{sec:implications}).
\end{itemize}

%% file: sections/questions.tex
\section{Evaluation Questions}
\label{sec:questions}
\label{sec:protocol}
We treat success as a property of a model evaluated under specified conditions. For a model $m$ and
evaluation conditions $c$,
\begin{equation}
S(m,c)=\E\left[\,Y \mid m,c\,\right], \qquad Y\in\{0,1\},
\label{eq:score}
\end{equation}
where $Y$ is the success of one episode, and the expectation is over the initial states and rollout
randomness specified by the study. The conditions $c$ encompass the initial states, the delivered instruction, and the execution horizon. A
model and the evaluation conditions together determine the policy that acts in closed loop. We hold the model fixed and vary the conditions. We define the
\emph{standard score} as the success rate on the initial states and instructions that the benchmark
provides at a stated execution horizon, which is the metric that benchmarks report. Other
interface choices, such as the observation format, the action parameterization, the episode time limit,
and the success check, stay fixed while a condition changes.

We ask one question for each evaluation choice, focusing on how a fixed model responds when exactly one
condition changes. For (1)~\emph{initial-state coverage}, we ask what performance variations the standard
initial states conceal. Section~\ref{sec:support} translates the target object while holding the instruction and execution
fixed. To select placements with evidence that the task can be solved there, it retains those on which a separate VLA model, used
as a reference, succeeds at a high rate (termed reference-selected conditions, Appendix~\ref{app:tau-sensitivity}). The task is unchanged at these placements. Appendix~\ref{app:theory} gives a minimal construction in which
success on the evaluated initial states does not determine success at such placements. For (2)~the \emph{delivered instruction}, we ask how much task-irrelevant text alters measured success. Section~\ref{sec:instruction}
removes or appends text while holding the scene and execution fixed. For (3)~\emph{action execution}, we ask
how the number of actions executed between observations changes both the standard score and
reference-selected success. Section~\ref{sec:execution} varies this horizon while holding the
instruction and initial states fixed. When two settings of a condition use the same initial states,
such as two execution horizons, episodes are paired by initial state.

Beyond these three diagnostics, we use two further designs. First, the adaptation case study uses the conditions of the state study to compare a base model with
its fine-tuned models under identical initial states and execution horizons, so the model changes while
the conditions remain fixed. Second, an instruction-following control on LIBERO-Goal tests whether a
model follows the instruction it receives: each episode is given the instruction of a different task,
and we record completion of both the requested and the original tasks. Throughout our analysis, a
\emph{response} denotes the change in a single model's success when a condition shifts, whereas a
\emph{comparison} denotes the performance margin between two models. We first quantify each model's
response, then evaluate in Section~\ref{sec:comparison} how these responses alter reported margins
between models.

%% file: sections/setup.tex
\section{Experimental Setup}
\label{sec:setup}
\noindent\textbf{Benchmarks and their roles.} We employ three benchmarks to isolate specific evaluation
conditions. From LIBERO~\citep{liu2023libero} we use four suites of ten tasks, Spatial, Object, Goal and Long (LIBERO-10), where each task is
evaluated from a fixed set of 50 initial states. We call LIBERO with the initial states and instructions
it provides \emph{standard LIBERO}, to distinguish it from the perturbed variants below. Within standard
LIBERO, we assign specific suites to distinct studies. We use LIBERO-Object, whose tasks require
picking a named object, to translate the target object and to run the suffix test (which appends neutral words, ordinary English words that name
no object, direction or action). We use LIBERO-Goal, whose ten tasks share a
single scene, for the instruction-following control, and all four suites to sweep over execution
horizons. LIBERO-Plus~\citep{fei2025liberoplus} provides 10,030 perturbed variants of the LIBERO tasks, with perturbations such as camera and lighting changes. We use it across all four suites to remove the perturbation
parameters appended to most instructions (the repair). Finally, we repeat the execution sweep in SimplerEnv~\citep{li2024simpler}, which evaluates released models in another simulator with the WidowX and Google Robot embodiments.

\Needspace{16\baselineskip}
\begin{wraptable}[15]{r}{0.67\linewidth}
\vspace{-\baselineskip}
\centering\footnotesize\setlength{\tabcolsep}{2.5pt}
\caption{\textbf{Models and the studies that evaluate them.}}
\label{tab:roles}
\vspace{2pt}%
\input{tables/roles}
\end{wraptable}
\noindent\textbf{Models and study coverage.} On LIBERO and LIBERO-Plus, we evaluate \NumPolicies{} VLA models spanning
\NumFamilies{}
families~\citep{black2024pi0,physicalintelligence2025pi05,shukor2025smolvla,sun2026vlajepa,fang2026molmoact2,lin2025evo1,zheng2025xvla,kim2025oft}.
Table~\ref{tab:roles} lists the models included in each study, grouped by condition. The state study evaluates five models, with a sixth, VLA-JEPA ($\circ$), as the reference. Five models ($\piz$, $\pif$, SmolVLA, SmolVLA-LP, and
X-VLA) are evaluated in all three fixed-model studies, enabling cross-study comparisons. The suffix ``-LP'' marks a model fine-tuned on LIBERO-Plus data, while the \emph{Actions} column specifies whether a model predicts delta
or absolute end-effector targets. Unless explicitly varied, each model uses the configuration distributed with it, including its default execution horizon $h$
(Table~\ref{tab:roles}), which ranges from \DefaultHorizonMin{} to \DefaultHorizonMax{}.

\noindent\textbf{Outcomes and uncertainty.} We report success rates in percent and performance changes
in percentage points (pp). The state study evaluates each LIBERO-Object task with the target object at
its original position and at five translated positions. Each position is evaluated from 20 initial
states under three different evaluation seeds, and all tasks are weighted equally. Standard-LIBERO evaluations run each
task's 50 initial states once, so that compared settings share their initial states
(Appendix~\ref{app:protocol}). The instruction study runs the same LIBERO-Plus variants under both
original and repaired instructions. Most confidence intervals resample the primary unit of variation of each design~\citep{owen2012bootstrap}. Appendices~\ref{app:protocol} and~\ref{app:stats}
give the sample sizes, estimands, and resampling units.

%% file: tables/roles.tex
\begin{tabular}{@{}llcccccc@{}}
\toprule
\multirow{2}{*}{Model} & \multirow{2}{*}{Actions} & \multirow{2}{*}{$h$} & \multicolumn{2}{c}{(1)} & \multicolumn{2}{c}{(2)} & (3)\\
\cmidrule(lr){4-5}\cmidrule(lr){6-7}\cmidrule(l){8-8}
 & & & States & Adapt. & Repair & Suffix & Horizon\\
\midrule
$\piz$ & delta & 50 & $\checkmark$ &  & $\checkmark$ & $\checkmark$ & $\checkmark$ \\
$\pif$ & delta & 50 & $\checkmark$ & $\checkmark$ & $\checkmark$ & $\checkmark$ & $\checkmark$ \\
SmolVLA & delta & 50 & $\checkmark$ & $\checkmark$ & $\checkmark$ & $\checkmark$ & $\checkmark$ \\
SmolVLA-LP & delta & 50 & $\checkmark$ &  & $\checkmark$ & $\checkmark$ & $\checkmark$ \\
EVO-1 & delta & 50 &  &  & $\checkmark$ & $\checkmark$ & $\checkmark$ \\
X-VLA & absolute & 30 & $\checkmark$ &  & $\checkmark$ & $\checkmark$ & $\checkmark$ \\
MolmoAct2 & delta & 10 &  &  & $\checkmark$ & $\checkmark$ &  \\
OpenVLA-OFT & delta & 8 &  &  & $\checkmark$ &  &  \\
OpenVLA-OFT-LP & delta & 8 &  &  & $\checkmark$ &  &  \\
VLA-JEPA & delta & 7 & $\circ$ &  & $\checkmark$ & $\checkmark$ &  \\
\bottomrule
\end{tabular}

%% file: sections/results.tex
\section{Findings}
\label{sec:findings}
\label{sec:results}

\subsection{Initial-state coverage}
\label{sec:support}
The standard initial states cover a narrow range of target placements. In \StatePinned{} of the
\StateTasks{} LIBERO-Object tasks, the target never moves across the 50 initial states, and in the
others it varies by at most \StateMaxTargetRangeCM{}\,cm. These positions also lie within
\DemoHausdorffMM{}\,mm of the starts of the training demonstrations distributed with LIBERO, and within
\FixedHausdorffMM{}\,mm where the target never moves (Appendix~\ref{app:states}). We test which
performance differences an evaluation on these states leaves unmeasured.

\begin{figure}[t]
\centering
\pendinggraphics[width=\linewidth]{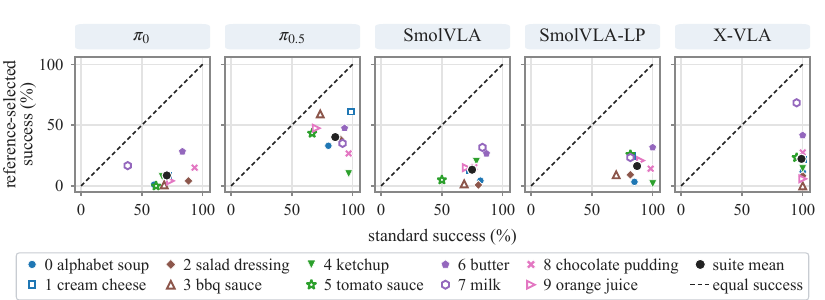}
\caption{\textbf{At default settings, every model scores lower on reference-selected conditions.} Each point is one LIBERO-Object task and black points are suite means, at each model's default horizon.}
\label{fig:support}
\end{figure}

\Needspace{12\baselineskip}
\begin{wraptable}[12]{r}{0.60\linewidth}
\vspace{-\baselineskip}
\centering\small\setlength{\tabcolsep}{2.6pt}
\caption{\textbf{State study at the default $h$ and at $h=10$.} {At LIBERO's 20\,Hz, $h$ actions last $h/20$\,s.}}
\label{tab:statehorizon}
\pendingsized{\input{tables/state_horizon_main}}
\vspace{-0.5\baselineskip}
\end{wraptable}
\vspace{-\parskip}\noindent\textbf{Evaluation protocol.} We translate the target by 4 to 12\,cm, outside the standard initial
states and the benchmark's specified initialization region, and leave the rest
of each standard initial state unchanged. Each task--translation
condition uses the same 20 initial states and three evaluation seeds. We keep as reference-selected
conditions the \RefKept{} of the 50 conditions on which VLA-JEPA, the reference, succeeds at least 75\% of the time.
The reference's success rate on a
condition is evidence that the condition can be solved, not that every initial state in it can be. For each of the five evaluated models, we report the standard score $B$, the reference-selected success $\Fref$
(averaged over each task's retained conditions, then equally over tasks), and the gap $B-\Fref$.
We evaluate $B$ and $\Fref$ at $h=10$ for all five models, using one fixed set of conditions selected by
the reference at $h=7$, and report each model at its default horizon on the same conditions
(Table~\ref{tab:statehorizon}). With each of the other evaluated models as the reference in turn, excluded from its
own comparison, or with thresholds of 50\% and 90\%, no model--task pair succeeds more often on the
reference-selected conditions than on the standard initial states (Appendix~\ref{app:tau-sensitivity}).

\Needspace{11\baselineskip}
\begin{wrapfigure}[10]{r}{0.42\linewidth}
\vspace{-\baselineskip}
\centering
\includegraphics[width=\linewidth]{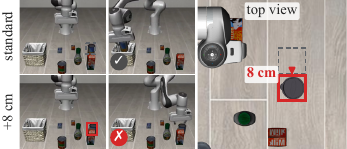}
\caption{\textbf{One task, two initial states.}}
\label{fig:rollouts}
\end{wrapfigure}
\noindent\textbf{Standard scores mask differences across placements.} At $h=10$, X-VLA
succeeds on \HTenXvlaB{}\% of the standard initial states and \HTenXvlaF{}\% of the reference-selected
conditions, whereas $\pif$ succeeds on \HTenPiFiveB{}\% and \HTenPiFiveF{}\%. Similar standard scores thus accompany a \HTenFieldGapPts{}-point difference between the two models on
the reference-selected conditions. Intervals that repeat the reference's selection in every bootstrap resample exclude zero for this margin and for every other margin between the five models at $h=10$. On the \HeldKept{} of the \HeldCands{} held-out conditions that the reference keeps (none among the original candidates), $\pif$ and X-VLA differ by \HeldPiXvla{}\,pp [\HeldPiXvlaLo{}, \HeldPiXvlaHi{}] (simultaneous 95\% interval, Appendix~\ref{app:tau-sensitivity}). Reference-selected
success is lower in \HTenLower{} of the \SupRankedRows{} model--task pairs and equal in one, so the gap
is not confined to a few tasks. At each model's default horizon, reference-selected success is lower in all \SupRankedRows{}
pairs (Figure~\ref{fig:support}). Figure~\ref{fig:rollouts} shows one task in which moving the target by
8\,cm changes X-VLA's outcome from success on all evaluated initial states to failure on all of them.

\begin{figure}[t]
\centering
\pendinggraphics[width=\linewidth]{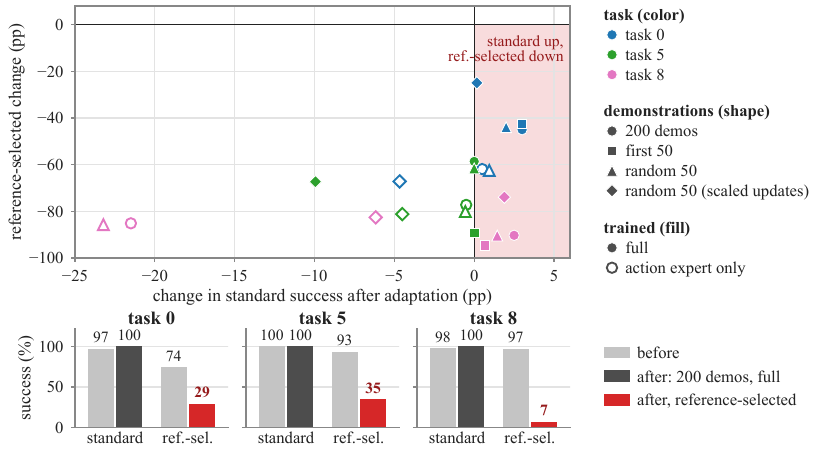}
\caption{\textbf{Adapting $\pif$: every regime loses reference-selected success.} Top: one point per task and regime, averaged over training runs. Bottom: full-model fine-tuning on 200 demonstrations. All at $h=5$.}
\label{fig:adaptation}
\end{figure}

\noindent\textbf{Fine-tuning can preserve standard success while losing success elsewhere.} We next apply these evaluations before and after task-specific fine-tuning of $\pif$, whose public checkpoint was already
fine-tuned on LIBERO. On tasks 0, 5 and 8, seven fine-tuning regimes vary the scripted demonstration set, the
trained parameters and the update budget, with three training runs per demonstration set
(Appendix~\ref{app:adaptation}). Base and fine-tuned models use $h=5$ on the same standard and reference-selected conditions. The evaluated standard initial states also occur in the
recorded demonstrations, and no demonstration begins at a translated placement. On task~\ExTask{},
standard success stays at \ExBenchTo{}\%, while reference-selected success falls from \ExFieldFrom{}\% to
\ExFieldTo{}\% (Figure~\ref{fig:adaptation}). All \CexContrasts{} task--regime settings reduce
reference-selected success, with simultaneous one-sided 95\% upper bounds below zero. In \BlindWideN{} of
them, the 90\% interval of the change in the standard score lies within $\pm$10\,pp
(Table~\ref{tab:adaptation-full}).

\noindent\textbf{Losses persist with fixed normalization.} The fine-tuning procedure used here recomputes the
normalization statistics of states and actions on the demonstration data, so it changes these statistics
along with the weights. To test whether this change explains the loss, we repeat full-model fine-tuning with the base statistics fixed
during training and evaluation. At $h=5$, reference-selected success still falls by \NcLossMin{} to
\NcLossMax{}\,pp, tasks 0 and 5 have
\NcTaskZeroB{}\% standard success, and that of task 8 falls to \NcTaskEightB{}\%. The loss remains at $h=10$. Fine-tuning $\pif$ on the benchmark's own human demonstrations of each of the ten tasks leaves the statistics unchanged and also lowers its reference-selected success, by \HumPiSel{}\,pp on average at $h=5$, while its standard score changes by $\HumPiStd$\,pp (Appendix~\ref{app:adaptation}). A separate control without
any update shows that changing the statistics alone can also reduce success
(Table~\ref{tab:zero-update}). Changing the statistics can
add to the loss, but a loss remains when they are held fixed.

\takeaway{(1) Initial-state coverage}{Standard success can stay high while success on reference-selected conditions is low or falls after fine-tuning. Evaluating both reveals differences and losses that the standard score can miss.}

\subsection{Delivered instruction}
\label{sec:instruction}
We next hold the scene fixed and test whether task-irrelevant text changes success, first in benchmark-generated instructions and then in controlled text variations.

\noindent\textbf{Evaluation protocol.} LIBERO-Plus appends perturbation parameters to the
instructions of \CensusMeta{} of its 10,030 variants (\CensusPct{}\%). The model therefore receives, for
example, \emph{pick up the alphabet soup and place it in the basket} followed by
\texttt{view 0 0 100 0 0 initstate 381}. This construction and its cost for one model were reported earlier in the benchmark's issue
tracker\footref{fn:issues}. We compare the
original instructions with repaired ones that remove these parameters, keeping the scene, the model and
its execution unchanged. Variants that rewrite the language are left unchanged
(Appendix~\ref{app:census}).

\Needspace{16\baselineskip}
\begin{wrapfigure}[16]{r}{0.5\linewidth}
\vspace{-\baselineskip}
\centering
\includegraphics[width=\linewidth]{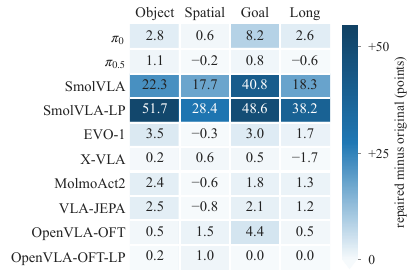}
\caption{\textbf{The largest repair effects occur in the SmolVLA pair.}}
\label{fig:instruction}
\end{wrapfigure}
\noindent\textbf{Repair effects differ across models.} Removing the appended parameters improves SmolVLA by
\InstrSmolMin{} to \InstrSmolMax{}\,pp and SmolVLA-LP by \InstrSmolPlusMin{} to \InstrSmolPlusMax{}\,pp
across the four suites (Figure~\ref{fig:instruction}). The other eight models change by at most
\InstrOtherMax{}\,pp. The large effect is specific to the two SmolVLA models, so comparisons that include either of them depend on the text the benchmark delivers. Across all Object variants, SmolVLA-LP, which was fine-tuned on LIBERO-Plus, improves
from \ObjectSmolPlusBefore{}\% to \ObjectSmolPlusAfter{}\%. Table~\ref{tab:instruction-intervals} reports
all rates and intervals. A seeded replication on 100 variants per suite reproduces all \PRCriterionTested{} repair effects above 5\,pp for $\piz$, $\pif$, SmolVLA and SmolVLA-LP (Appendix~\ref{app:pairedrng}). Over all ten models at their default horizons, it finds the largest effects in the SmolVLA pair, which gain \PRSmolAll{} and \PRSmolPlusAll{}\,pp pooled over the suites, while no other model changes by more than \PROthersSuiteMax{}\,pp on any suite (Table~\ref{tab:seeded-all}).

\noindent\textbf{Filler text reproduces large losses.} To test whether the appended parameters
themselves are required, we pad repaired instructions to their original word count with neutral filler, on a fixed 100-variant subset of each suite. On \LenMainSuite{}, SmolVLA-LP succeeds on 10.0\% of
the subset with the original text, 62.3\% after the repair and 11.3\% after the padding
(Table~\ref{tab:length-matched}). The filler lowers success in all \LenCells{} SmolVLA model--suite
pairs, so the appended parameters are not required for a large loss.

\Needspace{14\baselineskip}
\begin{wrapfigure}[14]{r}{0.44\linewidth}
\vspace{-\baselineskip}
\centering
\pendinggraphics[width=\linewidth]{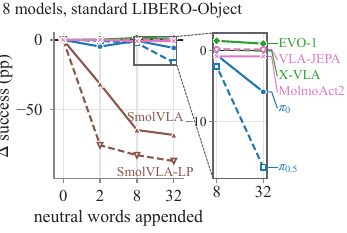}
\caption{\textbf{Neutral words after the benchmark's own instruction.}}
\label{fig:suffix}
\end{wrapfigure}
Sensitivity also appears on standard LIBERO-Object. Appending two neutral words lowers SmolVLA-LP by \SufWorstTwoLoss{}\,pp and SmolVLA by \SufSmolTwoLoss{}\,pp. Two words drawn at random from 200 ordinary words lower SmolVLA-LP and SmolVLA by similar amounts, \RWSmolPlusPooled{} and \RWSmolPooled{}\,pp averaged over three independent draws, while each of the other six models changes by at most \RWOthersMax{}\,pp on average (Table~\ref{tab:random-words}). With the
neutral suffix lengthened to \SufWorstK{} words, SmolVLA-LP and SmolVLA fall to \SufWorstTo{}\% and \SufSmolTo{}\%, $\pif$ loses
\SufPiFiveLongLoss{}\,pp, and the other five models stay within \SufSmallMax{}\,pp at every length (Figure~\ref{fig:suffix}, Appendix~\ref{app:suffix}).

\noindent\textbf{Following instructions does not ensure robustness to task-irrelevant text.} LIBERO-Goal
places its ten tasks in one shared scene. We give each episode the instruction of another task and record
completion of both the requested and the original task. Every tested model completes the requested task
in \GoalInstructedMin{} to \GoalInstructedMax{}\% of episodes and the original task in none. Yet two neutral words appended to a task's own instruction change the success of SmolVLA and SmolVLA-LP by
$\GoalSmolVLASfxTwoDelta$ and $\GoalSmolVLAPlusSfxTwoDelta$\,pp (Table~\ref{tab:goal-instruction}).

\takeaway{(2) Delivered instruction}{Task-irrelevant text can change success even when a model follows
task-changing instructions. Robustness evaluations should therefore control the text delivered alongside
the scene.}

\subsection{Action execution}
\label{sec:execution}
Finally, we vary $h$, the number of predicted actions executed before the next observation, and measure
the response on both standard and reference-selected conditions.

\noindent\textbf{Evaluation protocol.} The six models of the execution study are evaluated at each
$h\in\{5,10,25\}$ that does not exceed the length of their predicted action sequence and at their default. The models,
instructions, action units and episode budgets stay fixed. We report the full grid and use $h=10$ for the comparison across models (Appendix~\ref{app:horizon}).

\begin{figure}[t]
\centering
\pendinggraphics[width=\linewidth]{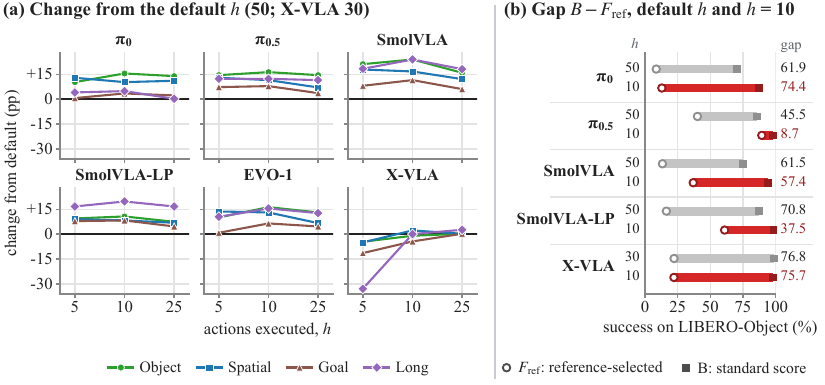}
\caption{\textbf{The execution horizon affects each model differently, and changes the gap.} (a)~Six models on four LIBERO suites. (b)~Five models on LIBERO-Object, both sets of initial states.}
\label{fig:execution}
\end{figure}

\noindent\textbf{Execution effects differ across models.} Relative to each model's default, $h=10$
improves five of the six models on all four LIBERO suites, by $\VanUpMin{}$ to $\VanDeltaMax{}$\,pp,
while the success of X-VLA changes by $\VanXvlaMin{}$ to $\VanXvlaMax{}$\,pp (Figure~\ref{fig:execution}a). Reducing
$h$ from 50 to 5 or 25 also raises the point estimate in all \NShortPairs{} model--suite pairs of the
five models whose default is 50. Within each of these pairs, success at $h=5$, 10 and 25 differs by at most 8.0\,pp. The direction of the response is not shared across models: on LIBERO-Long, $h=5$ raises the success of these five models by 4.0 to 18.2\,pp over their defaults and lowers that of X-VLA, whose default is 30, from 91.0\% to 58.2\% (Table~\ref{tab:hcurve}).

\noindent\textbf{Execution also changes the gap.} On the fixed reference-selected conditions,
$h=10$ raises success for \GapFieldUp{} of the \GapPanel{} state-study models, and every gap keeps its
sign. The gap narrows by \GapPiFiveChangePP{}\,pp for $\pif$, widens by \GapPiZeroChangePP{}\,pp for $\piz$, whose standard success rises more than its reference-selected success, and changes little for X-VLA (Figure~\ref{fig:execution}b, Table~\ref{tab:statehorizon}). The difference between standard and
reference-selected success therefore depends on execution as well as on placement.

\noindent\textbf{Execution responses differ across robot configurations.} We repeat the sweep in
SimplerEnv~\citep{li2024simpler} with the X-VLA checkpoints trained separately for WidowX and for Google
Robot~\citep{zheng2025xvla} and with DB-CogACT, the CogACT model of the
Dexbotic toolbox, on WidowX~\citep{li2024cogact,xie2025dexbotic}. All three
configurations perform worst at their shortest tested horizon. X-VLA on WidowX rises from \SEbigLoPct{}\% at
$h=\SEbigLo{}$ to \SEbigHiPct{}\% at $h=\SEbigHi{}$, and DB-CogACT from \SECogLoPct{}\% at
$h=\SECogLo{}$ to \SECogHiPct{}\% at $h=\SECogHi{}$. X-VLA on Google Robot peaks at $h=10$, not at
$h=30$ (Table~\ref{tab:simplerenv-horizon}). Neither shorter nor longer execution is uniformly better
across these configurations.

\takeaway{(3) Action execution}{The execution horizon affects both standard success and the gap to
reference-selected success. Responses should be measured for the evaluated model and robot
configuration.}

\subsection{Consequences for model comparisons}
\label{sec:comparison}
Model-dependent responses change the margins between models, even when their order is preserved. For
models $i$ and $j$ under condition $c$, let $D_{ij}(c)=S(i,c)-S(j,c)$ be their margin, and let
$\delta_i=S(i,c_1)-S(i,c_0)$ be the response of model $i$ to a change from $c_0$ to $c_1$. Then
\begin{equation}
K_{ij}=D_{ij}(c_1)-D_{ij}(c_0)=\delta_i-\delta_j .
\label{eq:kij}
\end{equation}
We call an ordering reversed when the unadjusted 95\% intervals of the two margins exclude zero with
opposite signs. The intervals are computed by resampling LIBERO-Plus variants, and because variants derived from the same LIBERO task (their base task) share its difficulty, we also resample base tasks. On LIBERO-Plus Object at $h=50$, the margin of $\piz$
over SmolVLA-LP is \IHPiZeroSmolPlusFiftyOrig{}\,pp before the instruction repair and
\IHPiZeroSmolPlusFiftyRep{}\,pp after it. The order of this pair also reverses at $h=10$, and both
reversals hold when base tasks are resampled (Table~\ref{tab:instr-horizon-pairs}). In contrast, the same repair reverses the order of $\pif$ and SmolVLA-LP at $h=50$ only when variants are
resampled, and at $h=10$, $\pif$ still leads after the repair.

Most changes affect margins without reversing orders (Appendix Figure~\ref{fig:reversal}). The gaps of the state study reverse one suite-average ordering (Appendix~\ref{app:inversions}), and switching from $h=50$ to $h=10$ reverses \HEndSupAll{} of the 40 pair--suite combinations among the five models whose
default is 50 (Appendix~\ref{app:comparisons}). No common shift reproduces any of the three changes: after subtracting the shift that fits all models
best, the largest remaining deviation on Object is \ResCoreHorizon{} to \ResCoreSupportTen{}\,pp across the
three, using the state gaps at $h=10$ (Appendix~\ref{app:residuals}). Evaluation conditions therefore change the size of an advantage even when its direction holds.

%% file: tables/state_horizon_main.tex
\begin{tabular}{@{}lcrrrrrr@{}}
\toprule
\multirow{2}{*}{Model} & Default & \multicolumn{3}{c}{At default $h$} & \multicolumn{3}{c}{At $h=10$} \\
\cmidrule(lr){3-5}\cmidrule(l){6-8}
& $h$ & \multicolumn{1}{c}{$B$} & \multicolumn{1}{c}{$\Fref$} & \multicolumn{1}{c}{Gap} & \multicolumn{1}{c}{$B$} & \multicolumn{1}{c}{$\Fref$} & \multicolumn{1}{c@{}}{Gap} \\
\midrule
$\pi_0$ & 50 & 70.5 & 8.6 & 61.9 & 87.3 & 12.9 & 74.4\,\textcolor{myvermilion}{{\scriptsize$(+12.5)$}} \\
$\pi_{0.5}$ & 50 & 85.7 & 40.2 & 45.5 & 98.2 & 89.5 & 8.7\,\textcolor{myblue}{{\scriptsize$(-36.8)$}} \\
SmolVLA & 50 & 74.8 & 13.3 & 61.5 & 94.3 & 36.9 & 57.4\,\textcolor{myblue}{{\scriptsize$(-4.1)$}} \\
SmolVLA-LP & 50 & 87.2 & 16.4 & 70.8 & 98.5 & 61.0 & 37.5\,\textcolor{myblue}{{\scriptsize$(-33.3)$}} \\
X-VLA & 30 & 99.0 & 22.3 & 76.8 & 97.8 & 22.1 & 75.7\,\textcolor{myblue}{{\scriptsize$(-1.1)$}} \\
\bottomrule
\end{tabular}

%% file: sections/related.tex
\section{Related Work}
\noindent\textbf{Benchmark validity and coverage.} Relative performance depends on the
benchmark~\citep{dehghani2021lottery} and, for language models, on the prompt
format~\citep{sclar2024format,mizrahi2024multiprompt}. For VLAs, \citet{jiang2026benchmarking} show
that shortcuts and overfitting limit what high scores establish, and other benchmarks test manipulation
beyond one clean setting~\citep{wang2024vlatest,pumacay2024colosseum,morgan2026colosseum}. LIBERO-PRO reports near collapse
when object positions change~\citep{zhou2025liberopro} and finds that models with high LIBERO scores differ under displacement. Moving the target can make a task unsolvable, so we keep only placements on which a separate model succeeds. We also compare the models at one execution horizon and report intervals that include the selection step. LIBERO-Plus finds models largely
insensitive to language variations~\citep{fei2025liberoplus}, although its issue tracker reports that the parameters it appends to instructions lower the success of one model\footref{fn:issues}. We measure the effect of removing them for ten models and separate a text effect from a failure to follow instructions with length-matched filler and task-switch instructions. We also add an analysis of what the standard initial states cover and a before-and-after adaptation comparison.

\noindent\textbf{Action execution.} How many predicted actions to execute before replanning is an
established design choice, studied through prediction and execution
horizons~\citep{chi2023diffusionpolicy}, chunk size~\citep{zhao2023act}, and methods that overlap or
adapt replanning~\citep{black2025rtc,feng2026dvac,wang2026autohorizon,liang2026aac,liu2024bidirectional}.
SmolVLA ablates the number of executed steps on LIBERO~\citep{shukor2025smolvla}, and OpenVLA-OFT specifies the number executed in its evaluation~\citep{kim2025oft}. We compare models at one common horizon and measure the response across models, simulators and initial states.

\noindent\textbf{Evaluation practice and adaptation.} vla-eval standardizes inference across code
bases~\citep{choi2026vlaeval}, and implementation choices and finite samples affect reported
results~\citep{henderson2018matters,snyder2026beyondbinary,agarwal2021edge}. Fine-tuning can trade out-of-distribution
accuracy for in-distribution gains~\citep{kumar2022finetuning}, and in VLAs, training a new action
expert can degrade knowledge transferred from the vision-language model~\citep{driess2025knowledge}.
Factor-wise studies separate generalization to object position from other
factors~\citep{xie2024decomposing}. Imitation learning collects corrections away from the
demonstrations~\citep{ross2011dagger,laskey2017dart}, and reachability analysis and safe reinforcement
learning characterize states from which a target can be reached or a violation
avoided~\citep{bansal2017hjreachability,thananjeyan2021recoveryrl}. Our reference-selected conditions instead give empirical evidence that a task can be completed under them.

%% file: sections/discussion.tex
\section{Evaluating VLA Models Under Stated Conditions}
\label{sec:implications}
\noindent\textbf{What the diagnostics suggest.} The controls narrow the explanations without
identifying mechanisms. Losses after fine-tuning persist when the normalization statistics are kept
fixed. Padding the repaired instruction back to its original length brings most of the instruction loss back,
and the tested models follow task-changing instructions, yet some remain sensitive to task-irrelevant
additions. Changing $h$ substantially reduces $\pif$'s gap, whereas X-VLA's gap changes little between the tested settings.

\noindent\textbf{An evaluation procedure.} (a)~Evaluate the standard initial states and additional
initial states on which another model succeeds, and repeat the selection with each other model as the
reference, excluded from its own comparison (Section~\ref{sec:support}). (b)~Compare conditions within a model on the same initial states,
with intervals over the study's sampling unit (Section~\ref{sec:setup}). (c)~Report each model's response to
the execution horizon and to task-irrelevant text instead of a single rate (Sections~\ref{sec:instruction}
and~\ref{sec:execution}). (d)~Report the margin between two models under each stated condition, since most condition changes alter its size rather than its sign, and call an advantage
consistent across the tested conditions only when the intervals support its direction throughout (Section~\ref{sec:comparison}). Applied at $h=10$, the procedure separates models that the standard score does not: $\pif$ and X-VLA, with standard scores of \HTenPiFiveB{}\% and \HTenXvlaB{}\%, differ by \HTenFieldGapPts{}\,pp on the reference-selected conditions. Under simultaneous intervals, the reference-selected score orders all 10 pairs of models while the standard score orders \HeldStdResolved{}. These orders hold on held-out placements, and \RefMarginSameTen{} of the \RefMarginCellsTen{} margins keep the sign they have with VLA-JEPA at 75\% when other references and thresholds select the conditions, with sizes that depend on how many tasks the reference keeps (Appendix~\ref{app:tau-sensitivity}).

\noindent\fbox{\parbox{\dimexpr\linewidth-2\fboxsep-2\fboxrule\relax}{\textbf{Reporting card.} To support
interpretation and reproduction of a VLA success rate, report
\begin{enumerate*}[label=(\arabic*)]
\item the evaluated initial states, with identifiers and task weighting;
\item the instruction the model actually received;
\item the execution horizon, and the control rate (actions per second);
\item the action interface and the episode budget;
\item the unit of randomness and the rule that pairs episodes; and
\item the rule used to select checkpoints or settings.
\end{enumerate*}}}

\noindent\textbf{Scope and limitations.} Three boundaries delimit our claims, each with a next step.
First, the evidence comes from simulation. In SimplerEnv, reference-selected success
agrees with real-robot rankings as well as the standard score does, with the same Kendall's $\tau$
(Appendix~\ref{app:external}), so reference-selected conditions serve as a diagnostic ahead of real-robot tests. Second, the state and instruction studies cover the LIBERO family, and the translations move one object. Within these studies, the large instruction-repair effect is specific to the two SmolVLA models, whose comparisons with other models therefore depend on the delivered text (Section~\ref{sec:instruction}). Fine-tuning $\pif$ on the benchmark's own human demonstrations also lowers its success off the standard states, by less than fine-tuning on the scripted demonstrations does (Appendix~\ref{app:adaptation}). More benchmarks would show how far the effects extend. Third, the models in each comparison share a control frequency, so equal $h$ corresponds to equal simulated time. Their action parameterizations can differ, and a common $h$ spans different times across environments. We report the response to $h$ rather than a recommended
value, and leave trial allocation across conditions to future work.

The standard score remains a useful measure under a specified evaluation, and the state results are
instances of underspecification~\citep{damour2022underspecification}, in which models that agree on
the evaluated conditions diverge elsewhere. Success rates are not model properties: a rate describes a
model under the conditions that produced it.

%% file: appendix/a_details.tex
\section{Experimental Details}
\label{app:config}
In this section, we describe the models we evaluate, the evaluation protocol of each study, and the statistics we report.

\textbf{Models.} We evaluate the public checkpoints in Table~\ref{tab:checkpoints} on LIBERO and LIBERO-Plus, each with the inference code and configuration it is distributed with; Appendix~\ref{app:simplerenv} describes the SimplerEnv models. The four suite-specific OpenVLA-OFT checkpoints, with the suffixes \texttt{spatial}, \texttt{object}, \texttt{goal} and \texttt{10}, are each evaluated on their own suite. X-VLA predicts absolute end-effector targets, and the other models predict changes in the end-effector pose. Table~\ref{tab:roles} lists the studies in which each model is evaluated. The code that moves objects for the state study does not support EVO-1 and MolmoAct2. The execution study leaves out VLA-JEPA, which predicts seven actions per call, and MolmoAct2, whose default already equals ten. The two OpenVLA-OFT variants run only through their own evaluation code, so they are evaluated in the instruction repair on LIBERO-Plus but not in the suffix test, the LIBERO-Goal control, or the state or execution studies.

\begin{table}[!htbp]
\centering\small
\caption{\textbf{Evaluated checkpoints.}}
\label{tab:checkpoints}
\vspace{4pt}%
\begin{tabular}{@{}ll@{}}
\toprule
Model & Checkpoint \\
\midrule
$\piz$ & \texttt{lerobot/pi0\_libero\_finetuned\_v044} \\
$\pif$ & \texttt{lerobot/pi05\_libero\_finetuned\_v044} \\
SmolVLA & \texttt{lerobot/smolvla\_libero} \\
SmolVLA-LP & \texttt{lerobot/smolvla\_libero\_plus} \\
VLA-JEPA & \texttt{lerobot/VLA-JEPA-LIBERO} \\
MolmoAct2 & \texttt{lerobot/MolmoAct2-LIBERO-LeRobot} \\
EVO-1 & \texttt{zuoxingdong/evo1\_libero} \\
X-VLA & \texttt{lerobot/xvla-libero} \\
OpenVLA-OFT & \texttt{moojink/openvla-7b-oft-finetuned-libero-}\textit{suite} \\
OpenVLA-OFT-LP & \texttt{sii-research/openvla-7b-oft-finetuned-libero-plus} \\
\bottomrule
\end{tabular}
\end{table}

\subsection{Evaluation protocol}
\label{app:protocol}
\textbf{Standard LIBERO.} Each evaluated LIBERO suite has ten tasks, and each task has 50 standard initial states. We run every initial state once, in a fixed order, so all conditions of a model are evaluated on the same 500 initial states per suite. An episode ends when the task succeeds or when the suite's step limit is reached: 280 steps on Spatial and Object, 300 on Goal and 520 on Long, at LIBERO's default control frequency of 20\,Hz, which our code does not change. The execution study evaluates every model at 5, 10 and 25 executed actions and at its default, with 500 episodes per suite and setting.

\textbf{Initial-state study.} Each LIBERO-Object task is evaluated under six conditions: the standard initial states and five translations of the target object. Each condition runs standard initial states 0 to 19, each under three evaluation seeds. That is, each condition is evaluated on 60 episodes (20 initial states $\times$ 3 seeds), and each model on 3{,}600 episodes over the ten tasks. The adaptation study evaluates the base model and every fine-tuned model in the same way on tasks 0, 5 and 8 (on all ten tasks for the human demonstrations of Appendix~\ref{app:adaptation}), except that the standard initial states are run under ten evaluation seeds, with 200 episodes per model and task, to measure changes in the standard score more precisely.

\textbf{Instruction study.} Each of the 10{,}030 LIBERO-Plus variants is run once with its original instruction and once with the repaired one. The two runs of a variant share its scene and initial state, so 20{,}060 episodes are planned per model. Appendix~\ref{app:instruction-tables} gives the paired outcomes obtained for each model.

\textbf{SimplerEnv.} Each configuration runs \SEeps{} episodes per task at each execution horizon of Table~\ref{tab:simplerenv-horizon}. The WidowX configurations use the four WidowX tasks (carrot on plate, eggplant in basket, spoon on towel, stack cube), and the Google Robot configuration uses the six Google Robot tasks (close drawer, move near, open drawer, pick coke can, pick object, place in closed drawer). Success rates weight the tasks of a configuration equally.

\subsection{Statistics}
\label{app:stats}
We report success rates in percent and differences in percentage points. Unless stated otherwise, intervals are 95\% bootstrap percentile intervals~\citep{owen2012bootstrap}, and every difference is paired: conditions that share initial states or variants are resampled together.

\textbf{State study.} A condition's success on a task is the mean over its 20 initial states and three evaluation seeds, $\Fref$ averages the task's kept conditions, and suite means weight the ten tasks equally. The bootstrap resamples initial states within each task, so the three seeds of a state move together, and applies the same draw to the standard condition, every translated condition and both horizons (2{,}000 draws). The kept conditions are fixed by the reference and are not resampled, so these intervals are conditional on the selected set, unlike those of Appendix~\ref{app:tau-sensitivity}, which repeat the selection in every draw.

\textbf{Adaptation.} Each draw resamples the initial states once per evaluation campaign, shared by every condition and evaluation seed within it; the base model and the regimes of a task come from four campaigns, so a contrast across two campaigns is not paired over initial states. Each draw also resamples the three training runs and the random 50-demonstration subsets once for all regimes. The \CexContrasts{} contrasts form one family with studentized max-$T$ bounds (Appendix~\ref{app:adaptation}).

\textbf{LIBERO-Plus.} The bootstrap resamples variants with one weight vector shared by every condition, so both runs of a variant move together. Appendix~\ref{app:cluster} resamples base tasks instead. The variation between repeated runs of one variant is not included in the resulting intervals, and Appendix~\ref{app:pairedrng} measures it.

\textbf{Standard LIBERO and model comparisons.} The bootstrap resamples initial states within each task, shared by every model and setting. A margin between two models uses the same draw for both models and both conditions (2{,}000 draws).

%% file: appendix/b_states.tex
\section{Initial States}
\label{app:state}
\input{tables/appendix_b_secref_aligned_numbers}
This section describes how the standard initial states vary, the per-task results of Section~\ref{sec:support}, the choice of threshold and reference, and the adaptation study with its normalization controls.

\subsection{What the standard initial states vary}
\label{app:states}
Each LIBERO-Object task contains 50 standard initial states that place seven objects on a table. Table~\ref{tab:state-variation} shows, for each task, how many of the seven objects move across the 50 states and how far the target object moves. The target object is the one named in the task instruction. In \StatePinned{} of the \StateTasks{} tasks, the target object stays stationary, and no object varies in height.

Each task file defines a 5\,cm by 5\,cm region for the target, and the stored states cover at most 39\% of its width. The original LIBERO demonstrations start the target in the same place as the stored states: each starting position in one set is within \DemoHausdorffMM{}\,mm of the closest starting position in the other, and within \FixedHausdorffMM{}\,mm on the tasks in which the target stays stationary. The five translations move the target by $+4$, $+8$, $+12$ and $-8$\,cm along the world $x$ axis, and by $-8$\,cm along $y$ (written $+4x$, $+8x$, $+12x$, $-8x$ and $-8y$), and keep the rest of the stored state unchanged. Each translated position lies outside the declared region and \ProbeDemoMinCM{} to \ProbeDemoMaxCM{}\,cm from the closest demonstration start.

In tasks \ProbeUncleanTasks{}, shifting the target by $-8$\,cm along the $y$ axis places it against a neighboring object, and one of the two is pushed away when the simulator settles. At each model's default horizon, leaving out these conditions does not change the comparison: \ProbeCleanLower{} of the \ProbeCleanPairs{} model--task pairs still succeed less often on the reference-selected conditions, and no model's mean changes by more than \ProbeCleanMeanShift{} points.

\begin{table}[!htbp]
\centering\small
\caption{\textbf{Variation across the 50 standard initial states of each LIBERO-Object task.} The target range is the larger of the target's ranges along the world $x$ and $y$ axes in the stored states.}
\label{tab:state-variation}
\vspace{4pt}%
\input{tables/app_state_variation}
\end{table}

\subsection{Per-task results}
\label{app:support}
The reference keeps \RefKept{} of the 50 conditions: all five on task~0; $+4x$, $-8x$ and $-8y$ on task~1; $+4x$, $+8x$, $-8x$ and $-8y$ on tasks~2 and~8; $+4x$ and $-8x$ on tasks~3, 5 and~9; $+4x$ and $-8y$ on tasks~4 and~6; and $+4x$ on task~7. Table~\ref{tab:off-support-full} gives the per-task results on these conditions at each model's default horizon and at $h=10$, and the change in the gap between the two horizons, whose interval excludes zero for $\piz$, $\pif$ and SmolVLA-LP. At five executed actions, the setting of the LIBERO example in its developers' code, $\pif$ reaches \HFivePiFiveB{}\% standard and \HFivePiFiveF{}\% reference-selected success, close to its values at ten.

\begin{table}[!htbp]
\centering\small
\caption{\textbf{Per-task success at the standard initial states and on the reference-selected conditions.}}
\label{tab:off-support-full}
\label{tab:state-horizon-pertask}
\label{app:horizon-support}
\vspace{4pt}%
\pendingsized{\input{tables/app_state_tasks}}
\end{table}

\FloatBarrier
\subsection{Choice of threshold and reference}
\label{app:tau-sensitivity}
Section~\ref{sec:support} keeps a translated condition when VLA-JEPA, the reference, succeeds in at least 75\% of its 60 episodes. Translated conditions can be solved: on task~0, a scripted controller with access to the simulator state solves each of 21 translations of the target in all of its 15 episodes. At thresholds of 50\% and 90\% (Table~\ref{tab:tau-sensitivity}), the standard score exceeds reference-selected success in every model--task pair that keeps a condition, at the default horizons and at ten actions, except for ties at ten actions in which $\pif$ succeeds in every episode on both sets. At 90\%, tasks \TauNinetyDropped{} keep no conditions.

With each of the five evaluated models as the reference instead, at each of the three thresholds (50\%, 75\% and 90\%) and left out of its own comparison, a weaker reference keeps fewer conditions, and a task without a kept condition is left undefined rather than scored as zero. The standard score is the higher of the two in \RefSwapBelowDef{} of the \RefSwapRowsDef{} model--task pairs that keep a condition at the default horizons, where VLA-JEPA is also scored, and in \RefSwapBelowTen{} of the \RefSwapRowsTen{} at ten actions. In the others, the model succeeds in every episode on both sets.

\begin{table}[!htbp]
\centering\small
\caption{\textbf{Success on the reference-selected conditions at three thresholds.} Standard success averages the ten LIBERO-Object tasks; each threshold column averages the tasks that keep at least one condition (tasks \TauNinetyDropped{} keep none at 90\%). At $h=10$, the kept conditions are those selected at the default horizon, as in Table~\ref{tab:statehorizon}.}
\label{tab:tau-sensitivity}
\label{tab:secref}
\vspace{4pt}%
\pendingsized{\input{tables/app_threshold}}
\end{table}

Table~\ref{tab:reference-margins} gives the margins under the main selection, VLA-JEPA at 75\%. Their intervals resample the initial states of the reference and the compared models jointly and repeat the selection in every draw (\SelPropDraws{} draws), so they include the uncertainty of selecting conditions from finite reference rollouts. At ten actions, the margin between $\pif$ and X-VLA has the interval [\SelPropPiXvlaLo, \SelPropPiXvlaHi] points. Tables~\ref{tab:reference-margins-ten} and~\ref{tab:reference-margins-def} give every pair under every reference and threshold, with each reference excluded from the pairs that contain it. At ten actions, \RefMarginSameTen{} of the \RefMarginCellsTen{} alternative combinations of pair, reference and threshold keep the sign of the margin under the main selection, and in \RefMarginSameSigTen{} of them the interval also excludes zero in that direction. The two that change sign use $\piz$ as the reference at 50\%, where it keeps conditions on two tasks. Under every reference and threshold that keeps at least three tasks, $\pif$ minus X-VLA lies between \RefPiXvlaMinThree{} and \RefPiXvlaMaxThree{} points. At the default horizons, where three models keep no conditions at 75\% when used as the reference, \RefMarginSameDef{} of the \RefMarginCellsDef{} alternative combinations keep the sign.

\begin{table}[!htbp]
\centering\small
\caption{\textbf{Margins between models under the main selection.} Each entry gives the first model's reference-selected success minus the second's, in points, with VLA-JEPA as the reference at 75\%, and a 95\% interval that includes the selection step. ``Tasks kept'' counts the tasks on which the reference keeps a condition.}
\label{tab:reference-margins}
\vspace{4pt}%
\pendingsized{\input{tables/app_reference_margins}}
\end{table}
\FloatBarrier

A second set of VLA-JEPA rollouts on the same initial states, with three other evaluation seeds, keeps \FreshKept{} conditions at 75\%, including all \FreshKeptOrig{} of the original selection, and its per-condition success correlates with the original at $r=\FreshRateCorr{}$. Under this selection at ten actions, no margin between the five models changes sign, and $\pif$ minus X-VLA is \FreshPiXvla{} points [\FreshPiXvlaLo, \FreshPiXvlaHi].

The comparison also holds on placements outside the original candidates. Each task adds five translations of the target, by $(-4, 0)$, $(0, 4)$, $(0, 8)$, $(0, -4)$ and $(4, 4)$\,cm along the world $x$ and $y$ axes, and VLA-JEPA, run as in the main selection, keeps \HeldKept{} of these \HeldCands{} conditions at 75\%. Each compared model runs all of these translations at ten actions on the same 20 initial states and three seeds, 3{,}000 episodes per model. Table~\ref{tab:heldout} gives the margins with simultaneous 95\% intervals over the ten pairs, from the maximum absolute studentized deviation across pairs in each draw. The intervals exclude zero for \HeldStdResolved{} of the ten pairs on the standard initial states and for all ten on the original and on the fresh selection. On the held-out placements, all \HeldResolved{} intervals exclude zero, in the order of the reference-selected score. $\pif$ and X-VLA, $+0.3$ points apart on the standard initial states, differ by \HeldPiXvla{} points [\HeldPiXvlaLo, \HeldPiXvlaHi] on the held-out placements.

\begin{table}[!htbp]
\centering\small
\caption{\textbf{Margins between models under each reference and threshold, $h=10$.} Entries as in Table~\ref{tab:reference-margins}, without intervals. ``--'' marks a pair that contains the reference, or a reference that keeps no condition.}
\label{tab:reference-margins-ten}
\vspace{4pt}%
\pendingsized{\input{tables/app_reference_margins_ten}}
\end{table}

\begin{table}[!htbp]
\centering\small
\caption{\textbf{Margins between models under each reference and threshold, default $h$.} Entries as in Table~\ref{tab:reference-margins-ten}.}
\label{tab:reference-margins-def}
\vspace{4pt}%
\pendingsized{\input{tables/app_reference_margins_def}}
\end{table}

\begin{table}[!htbp]
\centering\small
\caption{\textbf{Margins between models on the standard initial states, on the selected conditions and on held-out placements.} Ten executed actions. Each entry is the first model's success minus the second's, in points, $\pm$ the half-width of a simultaneous 95\% interval over the ten pairs. The selected conditions come from VLA-JEPA's original rollouts or from its rollouts with three other seeds on the original candidates, and the held-out placements are the further translations that VLA-JEPA keeps.}
\label{tab:heldout}
\vspace{4pt}%
\pendingsized{\input{tables/app_heldout}}
\end{table}

\FloatBarrier
\subsection{Fine-tuning in the adaptation case study}
\label{app:adaptation}
The base model is the public $\pif$ LIBERO checkpoint, which was fine-tuned on all four LIBERO suites before this study. A scripted controller with access to the simulator state completes tasks 0, 1, 5, 6 and 8 reliably, and for each of tasks 0, 5 and 8 it records 200 demonstrations, four from each stored initial state. It moves the end effector and the gripper without rotation, and every demonstration uses the task's own instruction. Table~\ref{tab:finetune} lists the fine-tuning hyperparameters, adapted for a single GPU from LeRobot's published settings for the $\pif$ LIBERO checkpoint. The regimes differ only in the demonstrations, the trained parameters and, in two regimes, the number of optimizer steps.

The seven regimes use all 200 demonstrations, the first 50, or a random subset of 50, and they train either the whole model or only the action expert with the vision-language backbone frozen. The first 50 demonstrations start once from every stored state, and the three random subsets cover 35, 38 and 35 of the 50 states. Two regimes scale the number of updates with the number of demonstrations. Each regime is trained three times per demonstration set, so a random-subset regime has nine fine-tuned models.

Every regime recomputes the normalization statistics of actions and states on the task's 200 demonstrations. The demonstrations contain no rotation, so the recomputed standard deviation of the rotation actions is zero, which fixes the fine-tuned model's rotation output at zero. Each fine-tuned model is evaluated at its last checkpoint with five executed actions. The standard initial states 0 to 19 used in evaluation are also initial states of the recorded demonstrations (each random subset of 50 covers 15 of them), and no demonstration starts from a translated condition.

The \CexContrasts{} task--regime contrasts form one family. Each contrast is the change in reference-selected success from the base model, with a one-sided 95\% upper bound equal to the estimate plus $q$ times its bootstrap standard error (20{,}000 resamples of initial states and training runs). The critical value $q=\AdaptCritUpper{}$ is the $\lceil 0.95\,(R+1)\rceil$-th order statistic, over the $R$ resamples, of the largest one-sided deviation across the family, each divided by its contrast's bootstrap standard error. Every upper bound lies below zero, the largest at \LossWorstUpper{}\,pp. The 90\% interval of the change in the standard score lies within $\pm$10 points in \BlindWideN{} settings and within $\pm$5 points in \BlindN{} (Table~\ref{tab:adaptation-full}).

\begin{table}[!htbp]
\centering\small
\caption{\textbf{Fine-tuning hyperparameters of the adaptation case study.}}
\label{tab:finetune}
\vspace{4pt}%
\begin{tabular}{@{}ll@{}}
\toprule
Hyperparameter & Value \\
\midrule
Optimizer & AdamW \\
Learning rate & $2.5\times10^{-5}$, 1{,}000 warmup steps, then decay \\
Batch size & 16 \\
Precision & bfloat16 \\
Optimizer steps & 12{,}000 ($12{,}000\times n/200$ for $n$ demonstrations in the two scaled regimes) \\
Trained parameters & whole model (4.14B) or action expert (0.69B) \\
Demonstrations & 200 per task, or a subset of 50 \\
Training runs & 3 per demonstration set \\
\bottomrule
\end{tabular}
\end{table}

Table~\ref{tab:adaptation-full} gives every regime on the three tasks.

\begin{table}[!htbp]
\centering\small
\caption{\textbf{Adaptation regimes on tasks 0, 5 and 8.} The change is in reference-selected success relative to the base model, and ``Within'' gives the smaller band, $\pm$5 or $\pm$10 points, that contains the 90\% interval of the change in the standard score.}
\label{tab:adaptation-full}
\vspace{4pt}%
\pendingsized{\input{tables/app_adaptation}}
\end{table}

\begin{figure}[!htbp]
\centering
\includegraphics{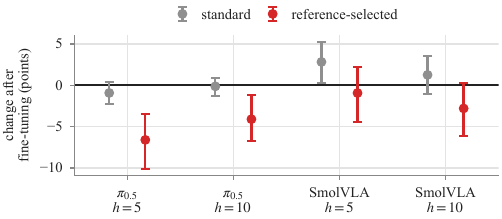}
\caption{{\textbf{Fine-tuning on the benchmark's human demonstrations, averaged over the ten tasks.} The change after fine-tuning in the standard score and in reference-selected success, in points, with 95\% intervals over initial states and training runs (Table~\ref{tab:human-ft}).}}
\label{fig:human-ft}
\end{figure}

\FloatBarrier
$\pif$ and SmolVLA are also fine-tuned in full on the benchmark's own human demonstrations of each of the ten LIBERO-Object tasks, three times per task. Each task has 42 to 50 such episodes in the LeRobot LIBERO dataset, and they include rotations. The base checkpoints were trained on the same dataset, so fine-tuning keeps their normalization statistics and the learning rate, schedule, batch size and number of steps of Table~\ref{tab:finetune}, $\pif$ in bfloat16 and SmolVLA in float32. Each fine-tuned model is evaluated on the initial states, conditions and seeds of its base model (Table~\ref{tab:human-ft} and Figure~\ref{fig:human-ft}). At five executed actions and averaged over the tasks, the standard score of $\pif$ changes by $\HumPiStd$ points [$\HumPiStdLo$, \HumPiStdHi], with a point estimate within 5 points of the base model on all \HumPiStdWithinFive{} tasks, and its reference-selected success falls by \HumPiSel{} points [\HumPiSelLo, \HumPiSelHi], or by \HumPiSelTen{} points [\HumPiSelTenLo, \HumPiSelTenHi] at ten actions. The standard score of SmolVLA rises by \HumSmolStd{} points [\HumSmolStdLo, \HumSmolStdHi] at five actions, and its reference-selected success changes by $\HumSmolSel$ points [$\HumSmolSelLo$, \HumSmolSelHi].

On tasks 0, 5 and 8, the reference-selected success of $\pif$ falls by \HumPiSelTaskA{}, \HumPiSelTaskF{} and \HumPiSelTaskI{} points after fine-tuning on the human demonstrations (Table~\ref{tab:human-ft}), and by \NcLossMin{} to \NcLossMax{} points after fine-tuning on the scripted ones with the same base statistics (Appendix~\ref{app:normalization}). The size of the loss therefore depends on the demonstrations and, with the same human demonstrations, on the model. SmolVLA fine-tuned in full on the 200 scripted demonstrations of each of the three tasks, three times per task with recomputed statistics (Table~\ref{tab:smolvla-scripted}), loses \SmolScrSelTaskA{}, \SmolScrSelTaskF{} and \SmolScrSelTaskI{} points of reference-selected success from \SmolBaseSelTaskA{}\%, \SmolBaseSelTaskF{}\% and \SmolBaseSelTaskI{}\% at five executed actions, and its standard score on task~0 falls by \SmolScrStdTaskA{} points from \SmolBaseStdTaskA{}\%.

\begin{table}[!htbp]
\centering\small
\caption{\textbf{Fine-tuning on the benchmark's human demonstrations.} Per task at five executed actions, the base model's success in percent and the change after fine-tuning in points, with 95\% intervals over initial states and training runs. The mean rows average the ten tasks at five and at ten executed actions.}
\label{tab:human-ft}
\vspace{4pt}%
\pendingsized{\input{tables/app_human_ft}}
\end{table}

\begin{table}[!htbp]
\centering\small
\caption{\textbf{SmolVLA fine-tuned on human and on scripted demonstrations.} The base model's success in percent and the change after fine-tuning in points, with 95\% intervals over initial states and training runs. The scripted demonstrations are those of Table~\ref{tab:adaptation-full}, with recomputed statistics.}
\label{tab:smolvla-scripted}
\vspace{4pt}%
\pendingsized{\input{tables/app_smolvla_scripted}}
\end{table}

\subsection{Normalization during fine-tuning}
\label{app:normalization}
Recomputing the normalization statistics changes what the model receives and emits before any update, so the loss after fine-tuning could come from the statistics rather than from the weight updates. Two controls separate the two.

First, the regime with all 200 demonstrations and the whole model is repeated with the base model's statistics kept during fine-tuning and evaluation (Tables~\ref{tab:normalization-control} and~\ref{tab:zero-update}). At five executed actions, reference-selected success still falls by \NcLossMin{} to \NcLossMax{} points, with every interval below zero, within \NcVsOrigMax{} points of the loss with recomputed statistics. At ten actions the loss is \NcLossTenMin{} to \NcLossTenMax{} points, again with every interval below zero. At five executed actions, the standard score of task~8 also falls, to \NcTaskEightB{}\%, so there the standard initial states show part of the loss.

Second, the base model is evaluated without any update, with the recomputed statistics swapped in (Table~\ref{tab:zero-update}). Swapping the input statistics alone changes success by amounts whose intervals include zero on all three tasks, and swapping both costs \NcZeroMin{} to \NcZeroMax{} points on all three tasks, mostly through the output statistics. The last row of Table~\ref{tab:zero-update} sets only the rotation output to zero, as the recomputed output statistics do. The statistics alone cost up to \NcZeroMax{} points, and the loss that remains with fixed statistics shows that the weight updates also reduce success on the reference-selected conditions.

\begin{table}[!htbp]
\centering\small
\caption{\textbf{Fine-tuning with the base model's normalization statistics.} Five executed actions; ``Recomputed'' is the procedure of Table~\ref{tab:adaptation-full}.}
\label{tab:normalization-control}
\vspace{4pt}%
\pendingsized{\input{tables/app_fixed_norm}}
\end{table}

\begin{table}[!htbp]
\centering\small
\caption{\textbf{Normalization controls as differences from the base model.} Change in reference-selected success, in points, with 95\% intervals, at five executed actions.}
\label{tab:zero-update}
\vspace{4pt}%
\pendingsized{\input{tables/app_norm_contrasts}}
\end{table}
\FloatBarrier

%% file: tables/appendix_b_secref_aligned_numbers.tex
\newcommand{\SecRefBelowAligned}{\pendingvalue{62}}
\newcommand{\SecRefRowsAligned}{\pendingvalue{68}}
\newcommand{\SecRefBaseRowsAligned}{\pendingvalue{40}}
\newcommand{\SecRefStarvedAligned}{\pendingvalue{7}}
\newcommand{\SecRefCountAligned}{\pendingvalue{4}}
\newcommand{\SecRefTiesAligned}{\pendingvalue{6}}

%% file: tables/app_state_variation.tex
\begin{tabular}{@{}lcc@{}}
\toprule
Task & Objects that move & Target range (cm) \\
\midrule
0 & 4 & 0.0 \\
1 & 4 & 1.0 \\
2 & 3 & 0.0 \\
3 & 3 & 1.9 \\
4 & 4 & 0.0 \\
5 & 4 & 0.0 \\
6 & 3 & 1.0 \\
7 & 4 & 1.0 \\
8 & 2 & 0.0 \\
9 & 3 & 0.0 \\
\bottomrule
\end{tabular}

%% file: tables/app_state_tasks.tex
\begin{tabular}{@{}lcrcrcrcrcr@{}}
\toprule
\multirow{2}{*}{Task} & \multicolumn{2}{c}{$\pi_0$} & \multicolumn{2}{c}{$\pi_{0.5}$} & \multicolumn{2}{c}{SmolVLA} & \multicolumn{2}{c}{SmolVLA-LP} & \multicolumn{2}{c}{X-VLA} \\
\cmidrule(lr){2-3}\cmidrule(lr){4-5}\cmidrule(lr){6-7}\cmidrule(lr){8-9}\cmidrule(lr){10-11}
 & Standard & \multicolumn{1}{c}{Ref.} & Standard & \multicolumn{1}{c}{Ref.} & Standard & \multicolumn{1}{c}{Ref.} & Standard & \multicolumn{1}{c}{Ref.} & Standard & \multicolumn{1}{c@{}}{Ref.} \\
\midrule
\multicolumn{11}{l}{\textit{Default executed actions}} \\
\midrule
0 & 60.0 & 1.0 & 80.0 & 33.0 & 81.7 & 4.3 & 85.0 & 3.3 & 100.0 & 11.7 \\
1 & 71.7 & 8.3 & 98.3 & 61.1 & 73.3 & 12.8 & 83.3 & 24.4 & 100.0 & 21.7 \\
2 & 88.3 & 4.2 & 90.0 & 38.3 & 80.0 & 0.8 & 81.7 & 9.2 & 100.0 & 8.3 \\
3 & 68.3 & 0.8 & 73.3 & 59.2 & 68.3 & 1.7 & 70.0 & 9.2 & 100.0 & 0.0 \\
4 & 66.7 & 7.5 & 96.7 & 10.0 & 78.3 & 20.0 & 100.0 & 1.7 & 100.0 & 14.2 \\
5 & 61.7 & 0.0 & 66.7 & 43.3 & 50.0 & 5.0 & 81.7 & 25.8 & 95.0 & 23.3 \\
6 & 83.3 & 28.3 & 93.3 & 47.5 & 86.7 & 26.7 & 100.0 & 31.7 & 100.0 & 41.7 \\
7 & 38.3 & 16.7 & 91.7 & 35.0 & 83.3 & 31.7 & 81.7 & 23.3 & 95.0 & 68.3 \\
8 & 93.3 & 15.0 & 96.7 & 26.7 & 76.7 & 15.4 & 98.3 & 14.2 & 100.0 & 27.5 \\
9 & 73.3 & 4.2 & 70.0 & 47.5 & 70.0 & 15.0 & 90.0 & 20.8 & 100.0 & 5.8 \\
\midrule
Mean & 70.5 & 8.6 & 85.7 & 40.2 & 74.8 & 13.3 & 87.2 & 16.4 & 99.0 & 22.3 \\
\midrule
\multicolumn{11}{l}{\textit{Ten executed actions}} \\
\midrule
0 & 80.0 & 2.7 & 100.0 & 82.3 & 91.7 & 21.7 & 96.7 & 44.3 & 100.0 & 14.7 \\
1 & 95.0 & 14.4 & 100.0 & 83.3 & 96.7 & 43.9 & 100.0 & 72.8 & 100.0 & 18.9 \\
2 & 98.3 & 11.3 & 95.0 & 82.9 & 95.0 & 11.7 & 100.0 & 45.0 & 100.0 & 10.8 \\
3 & 85.0 & 4.2 & 93.3 & 90.0 & 95.0 & 17.5 & 98.3 & 48.3 & 100.0 & 0.8 \\
4 & 88.3 & 5.0 & 98.3 & 75.0 & 93.3 & 30.0 & 98.3 & 25.0 & 100.0 & 11.7 \\
5 & 76.7 & 7.5 & 98.3 & 97.5 & 90.0 & 32.5 & 96.7 & 83.3 & 100.0 & 15.0 \\
6 & 93.3 & 35.8 & 100.0 & 100.0 & 98.3 & 76.7 & 98.3 & 88.3 & 100.0 & 45.0 \\
7 & 63.3 & 18.3 & 96.7 & 93.3 & 95.0 & 41.7 & 98.3 & 70.0 & 83.3 & 73.3 \\
8 & 100.0 & 16.7 & 100.0 & 99.6 & 90.0 & 52.1 & 100.0 & 73.8 & 100.0 & 26.3 \\
9 & 93.3 & 13.3 & 100.0 & 90.8 & 98.3 & 41.7 & 98.3 & 59.2 & 95.0 & 5.0 \\
\midrule
Mean & 87.3 & 12.9 & 98.2 & 89.5 & 94.3 & 36.9 & 98.5 & 61.0 & 97.8 & 22.1 \\
\midrule
\multicolumn{11}{l}{\textit{Change in the gap from the default to ten executed actions, with its 95\% interval}} \\
\midrule
 & \multicolumn{2}{c}{$+$12.5} & \multicolumn{2}{c}{$-$36.8} & \multicolumn{2}{c}{$-$4.1} & \multicolumn{2}{c}{$-$33.3} & \multicolumn{2}{c}{$-$1.1} \\
 & \multicolumn{2}{c}{{\footnotesize [$+$8.1, $+$17.0]}} & \multicolumn{2}{c}{{\footnotesize [$-$41.3, $-$32.4]}} & \multicolumn{2}{c}{{\footnotesize [$-$10.1, $+$2.3]}} & \multicolumn{2}{c}{{\footnotesize [$-$37.3, $-$29.2]}} & \multicolumn{2}{c}{{\footnotesize [$-$4.4, $+$2.0]}} \\
\bottomrule
\end{tabular}

%% file: tables/app_threshold.tex
\begin{tabular}{@{}lcrrrcrrr@{}}
\toprule
& \multicolumn{4}{c}{Default $h$} & \multicolumn{4}{c}{$h=10$} \\
\cmidrule(lr){2-5}\cmidrule(lr){6-9}
\multirow{2}{*}{Model} & \multirow{2}{*}{Standard} & \multicolumn{3}{c}{Threshold} & \multirow{2}{*}{Standard} & \multicolumn{3}{c}{Threshold} \\
\cmidrule(lr){3-5}\cmidrule(lr){7-9}
& & \multicolumn{1}{c}{50\%} & \multicolumn{1}{c}{75\%} & \multicolumn{1}{c}{90\%} & & \multicolumn{1}{c}{50\%} & \multicolumn{1}{c}{75\%} & \multicolumn{1}{c@{}}{90\%} \\
\midrule
$\pi_0$ & 70.5 & 6.4 & 8.6 & 12.3 & 87.3 & 10.3 & 12.9 & 19.0 \\
$\pi_{0.5}$ & 85.7 & 37.1 & 40.2 & 48.4 & 98.2 & 86.7 & 89.5 & 96.0 \\
SmolVLA & 74.8 & 10.5 & 13.3 & 17.7 & 94.3 & 31.3 & 36.9 & 49.6 \\
SmolVLA-LP & 87.2 & 14.8 & 16.4 & 21.8 & 98.5 & 56.3 & 61.0 & 74.4 \\
X-VLA & 99.0 & 16.0 & 22.3 & 35.0 & 97.8 & 15.7 & 22.1 & 36.1 \\
\midrule
Tasks with a kept condition & -- & 10 & 10 & 8 & -- & 10 & 10 & 8 \\
Pairs with standard above & -- & 50/50 & 50/50 & 40/40 & -- & 50/50 & 49/50 & 37/40 \\
\bottomrule
\end{tabular}

%% file: tables/app_reference_margins.tex
\begin{tabular}{@{}lcc@{}}
\toprule
Pair & $h=10$ & Default $h$ \\
\midrule
$\pi_0$ $-$ $\pi_{0.5}$ & $-$76.6 {\footnotesize [$-$79.8, $-$73.4]} & $-$31.6 {\footnotesize [$-$35.8, $-$27.2]} \\
$\pi_0$ $-$ SmolVLA & $-$24.0 {\footnotesize [$-$27.5, $-$19.5]} & $-$4.7 {\footnotesize [$-$7.7, $-$1.5]} \\
$\pi_0$ $-$ SmolVLA-LP & $-$48.1 {\footnotesize [$-$51.9, $-$43.5]} & $-$7.8 {\footnotesize [$-$10.7, $-$5.1]} \\
$\pi_0$ $-$ X-VLA & $-$9.2 {\footnotesize [$-$12.0, $-$4.2]} & $-$13.7 {\footnotesize [$-$16.3, $-$9.0]} \\
$\pi_{0.5}$ $-$ SmolVLA & $+$52.6 {\footnotesize [$+$48.6, $+$57.6]} & $+$26.8 {\footnotesize [$+$22.3, $+$31.6]} \\
$\pi_{0.5}$ $-$ SmolVLA-LP & $+$28.5 {\footnotesize [$+$23.9, $+$34.0]} & $+$23.8 {\footnotesize [$+$19.5, $+$28.1]} \\
$\pi_{0.5}$ $-$ X-VLA & $+$67.3 {\footnotesize [$+$63.8, $+$73.3]} & $+$17.9 {\footnotesize [$+$14.2, $+$24.0]} \\
SmolVLA $-$ SmolVLA-LP & $-$24.1 {\footnotesize [$-$29.6, $-$19.1]} & $-$3.0 {\footnotesize [$-$7.1, $+$0.8]} \\
SmolVLA $-$ X-VLA & $+$14.8 {\footnotesize [$+$10.1, $+$20.2]} & $-$8.9 {\footnotesize [$-$12.5, $-$3.8]} \\
SmolVLA-LP $-$ X-VLA & $+$38.9 {\footnotesize [$+$34.1, $+$45.7]} & $-$5.9 {\footnotesize [$-$8.9, $+$0.2]} \\
Tasks kept & 10 & 10 \\
\bottomrule
\end{tabular}

%% file: tables/app_reference_margins_ten.tex
\begin{tabular}{@{}lcccccc@{}}
\toprule
& \multicolumn{6}{c}{Reference} \\
\cmidrule(lr){2-7}
Pair & VLA-JEPA & $\pi_0$ & $\pi_{0.5}$ & SmolVLA & SmolVLA-LP & X-VLA \\
\midrule
\multicolumn{7}{@{}l}{\textit{$h=10$, threshold 50\%}} \\
$\pi_0$ $-$ $\pi_{0.5}$ & $-$76.4 & -- & -- & $-$73.2 & $-$74.2 & $-$55.8 \\
$\pi_0$ $-$ SmolVLA & $-$21.0 & -- & $-$16.7 & -- & $-$21.0 & $-$20.4 \\
$\pi_0$ $-$ SmolVLA-LP & $-$46.0 & -- & $-$40.0 & $-$49.7 & -- & $-$41.3 \\
$\pi_0$ $-$ X-VLA & $-$5.4 & -- & $-$3.6 & $-$12.6 & $-$6.2 & -- \\
$\pi_{0.5}$ $-$ SmolVLA & $+$55.4 & $+$20.0 & -- & -- & $+$53.2 & $+$35.4 \\
$\pi_{0.5}$ $-$ SmolVLA-LP & $+$30.4 & $+$7.5 & -- & $+$23.5 & -- & $+$14.6 \\
$\pi_{0.5}$ $-$ X-VLA & $+$71.0 & $+$5.0 & -- & $+$60.6 & $+$67.9 & -- \\
SmolVLA $-$ SmolVLA-LP & $-$25.0 & $-$12.5 & $-$23.3 & -- & -- & $-$20.8 \\
SmolVLA $-$ X-VLA & $+$15.6 & $-$15.0 & $+$13.1 & -- & $+$14.7 & -- \\
SmolVLA-LP $-$ X-VLA & $+$40.6 & $-$2.5 & $+$36.4 & $+$37.1 & -- & -- \\
Tasks kept & 10 & 2 & 10 & 6 & 10 & 4 \\
\midrule
\multicolumn{7}{@{}l}{\textit{$h=10$, threshold 75\%}} \\
$\pi_0$ $-$ $\pi_{0.5}$ & $-$76.6 & -- & -- & $-$71.9 & $-$75.6 & $-$30.8 \\
$\pi_0$ $-$ SmolVLA & $-$24.0 & -- & $-$22.4 & -- & $-$30.4 & $-$10.8 \\
$\pi_0$ $-$ SmolVLA-LP & $-$48.1 & -- & $-$41.0 & $-$65.6 & -- & $-$23.3 \\
$\pi_0$ $-$ X-VLA & $-$9.2 & -- & $-$4.9 & $-$5.8 & $-$8.4 & -- \\
$\pi_{0.5}$ $-$ SmolVLA & $+$52.6 & -- & -- & -- & $+$45.2 & $+$20.0 \\
$\pi_{0.5}$ $-$ SmolVLA-LP & $+$28.5 & -- & -- & $+$6.4 & -- & $+$7.5 \\
$\pi_{0.5}$ $-$ X-VLA & $+$67.3 & -- & -- & $+$66.1 & $+$67.2 & -- \\
SmolVLA $-$ SmolVLA-LP & $-$24.1 & -- & $-$18.5 & -- & -- & $-$12.5 \\
SmolVLA $-$ X-VLA & $+$14.8 & -- & $+$17.5 & -- & $+$22.0 & -- \\
SmolVLA-LP $-$ X-VLA & $+$38.9 & -- & $+$36.1 & $+$59.7 & -- & -- \\
Tasks kept & 10 & 0 & 10 & 3 & 8 & 2 \\
\midrule
\multicolumn{7}{@{}l}{\textit{$h=10$, threshold 90\%}} \\
$\pi_0$ $-$ $\pi_{0.5}$ & $-$77.1 & -- & -- & -- & $-$65.5 & $-$30.8 \\
$\pi_0$ $-$ SmolVLA & $-$30.6 & -- & $-$21.0 & -- & $-$18.7 & $-$10.8 \\
$\pi_0$ $-$ SmolVLA-LP & $-$55.4 & -- & $-$49.3 & -- & -- & $-$23.3 \\
$\pi_0$ $-$ X-VLA & $-$17.2 & -- & $-$11.3 & -- & $-$6.0 & -- \\
$\pi_{0.5}$ $-$ SmolVLA & $+$46.5 & -- & -- & -- & $+$46.8 & $+$20.0 \\
$\pi_{0.5}$ $-$ SmolVLA-LP & $+$21.7 & -- & -- & -- & -- & $+$7.5 \\
$\pi_{0.5}$ $-$ X-VLA & $+$59.9 & -- & -- & -- & $+$59.5 & -- \\
SmolVLA $-$ SmolVLA-LP & $-$24.8 & -- & $-$28.3 & -- & -- & $-$12.5 \\
SmolVLA $-$ X-VLA & $+$13.4 & -- & $+$9.7 & -- & $+$12.7 & -- \\
SmolVLA-LP $-$ X-VLA & $+$38.2 & -- & $+$38.0 & -- & -- & -- \\
Tasks kept & 8 & 0 & 9 & 0 & 5 & 2 \\
\bottomrule
\end{tabular}

%% file: tables/app_reference_margins_def.tex
\begin{tabular}{@{}lcccccc@{}}
\toprule
& \multicolumn{6}{c}{Reference} \\
\cmidrule(lr){2-7}
Pair & VLA-JEPA & $\pi_0$ & $\pi_{0.5}$ & SmolVLA & SmolVLA-LP & X-VLA \\
\midrule
\multicolumn{7}{@{}l}{\textit{Default $h$, threshold 50\%}} \\
$\pi_0$ $-$ $\pi_{0.5}$ & $-$30.7 & -- & -- & $-$20.0 & $-$20.0 & $-$32.9 \\
$\pi_0$ $-$ SmolVLA & $-$4.2 & -- & $-$2.3 & -- & $+$5.0 & $-$5.0 \\
$\pi_0$ $-$ SmolVLA-LP & $-$8.4 & -- & $-$12.5 & $-$1.7 & -- & $-$5.0 \\
$\pi_0$ $-$ X-VLA & $-$9.6 & -- & $-$20.6 & $-$33.3 & $-$33.3 & -- \\
$\pi_{0.5}$ $-$ SmolVLA & $+$26.6 & $+$25.0 & -- & -- & $+$25.0 & $+$27.9 \\
$\pi_{0.5}$ $-$ SmolVLA-LP & $+$22.3 & $+$18.3 & -- & $+$18.3 & -- & $+$27.9 \\
$\pi_{0.5}$ $-$ X-VLA & $+$21.1 & $-$13.3 & -- & $-$13.3 & $-$13.3 & -- \\
SmolVLA $-$ SmolVLA-LP & $-$4.2 & $-$6.7 & $-$10.2 & -- & -- & $+$0.0 \\
SmolVLA $-$ X-VLA & $-$5.4 & $-$38.3 & $-$18.3 & -- & $-$38.3 & -- \\
SmolVLA-LP $-$ X-VLA & $-$1.2 & $-$31.7 & $-$8.1 & $-$31.7 & -- & -- \\
Tasks kept & 10 & 2 & 7 & 2 & 2 & 4 \\
\midrule
\multicolumn{7}{@{}l}{\textit{Default $h$, threshold 75\%}} \\
$\pi_0$ $-$ $\pi_{0.5}$ & $-$31.6 & -- & -- & -- & -- & $-$20.0 \\
$\pi_0$ $-$ SmolVLA & $-$4.7 & -- & $-$8.8 & -- & -- & $+$5.0 \\
$\pi_0$ $-$ SmolVLA-LP & $-$7.8 & -- & $-$12.1 & -- & -- & $-$1.7 \\
$\pi_0$ $-$ X-VLA & $-$13.7 & -- & $-$25.4 & -- & -- & -- \\
$\pi_{0.5}$ $-$ SmolVLA & $+$26.8 & -- & -- & -- & -- & $+$25.0 \\
$\pi_{0.5}$ $-$ SmolVLA-LP & $+$23.8 & -- & -- & -- & -- & $+$18.3 \\
$\pi_{0.5}$ $-$ X-VLA & $+$17.9 & -- & -- & -- & -- & -- \\
SmolVLA $-$ SmolVLA-LP & $-$3.0 & -- & $-$3.3 & -- & -- & $-$6.7 \\
SmolVLA $-$ X-VLA & $-$8.9 & -- & $-$16.7 & -- & -- & -- \\
SmolVLA-LP $-$ X-VLA & $-$5.9 & -- & $-$13.3 & -- & -- & -- \\
Tasks kept & 10 & 0 & 4 & 0 & 0 & 2 \\
\midrule
\multicolumn{7}{@{}l}{\textit{Default $h$, threshold 90\%}} \\
$\pi_0$ $-$ $\pi_{0.5}$ & $-$36.1 & -- & -- & -- & -- & $-$8.3 \\
$\pi_0$ $-$ SmolVLA & $-$5.4 & -- & -- & -- & -- & $+$6.7 \\
$\pi_0$ $-$ SmolVLA-LP & $-$9.5 & -- & -- & -- & -- & $+$3.3 \\
$\pi_0$ $-$ X-VLA & $-$22.7 & -- & -- & -- & -- & -- \\
$\pi_{0.5}$ $-$ SmolVLA & $+$30.7 & -- & -- & -- & -- & $+$15.0 \\
$\pi_{0.5}$ $-$ SmolVLA-LP & $+$26.7 & -- & -- & -- & -- & $+$11.7 \\
$\pi_{0.5}$ $-$ X-VLA & $+$13.4 & -- & -- & -- & -- & -- \\
SmolVLA $-$ SmolVLA-LP & $-$4.1 & -- & -- & -- & -- & $-$3.3 \\
SmolVLA $-$ X-VLA & $-$17.3 & -- & -- & -- & -- & -- \\
SmolVLA-LP $-$ X-VLA & $-$13.2 & -- & -- & -- & -- & -- \\
Tasks kept & 8 & 0 & 0 & 0 & 0 & 1 \\
\bottomrule
\end{tabular}

%% file: tables/app_heldout.tex
\begin{tabular}{@{}lcccc@{}}
\toprule
& Standard & \multicolumn{2}{c}{Selected conditions} & Held-out \\
\cmidrule(lr){3-4}
Pair & initial states & original rollouts & fresh rollouts & placements \\
\midrule
$\pi_0$ $-$ $\pi_{0.5}$ & $-$10.8 {\footnotesize $\pm$3.8} & $-$76.6 {\footnotesize $\pm$4.6} & $-$76.5 {\footnotesize $\pm$4.6} & $-$75.2 {\footnotesize $\pm$2.8} \\
$\pi_0$ $-$ SmolVLA & $-$7.0 {\footnotesize $\pm$5.1} & $-$24.0 {\footnotesize $\pm$5.7} & $-$23.2 {\footnotesize $\pm$5.8} & $-$17.3 {\footnotesize $\pm$3.7} \\
$\pi_0$ $-$ SmolVLA-LP & $-$11.2 {\footnotesize $\pm$3.8} & $-$48.1 {\footnotesize $\pm$6.2} & $-$45.6 {\footnotesize $\pm$5.6} & $-$45.9 {\footnotesize $\pm$3.5} \\
$\pi_0$ $-$ X-VLA & $-$10.5 {\footnotesize $\pm$3.7} & $-$9.2 {\footnotesize $\pm$5.7} & $-$9.0 {\footnotesize $\pm$4.9} & $-$7.5 {\footnotesize $\pm$3.3} \\
$\pi_{0.5}$ $-$ SmolVLA & $+$3.8 {\footnotesize $\pm$3.9} & $+$52.6 {\footnotesize $\pm$6.4} & $+$53.3 {\footnotesize $\pm$6.5} & $+$57.9 {\footnotesize $\pm$3.3} \\
$\pi_{0.5}$ $-$ SmolVLA-LP & $-$0.3 {\footnotesize $\pm$2.0} & $+$28.5 {\footnotesize $\pm$7.3} & $+$31.0 {\footnotesize $\pm$6.5} & $+$29.2 {\footnotesize $\pm$3.5} \\
$\pi_{0.5}$ $-$ X-VLA & $+$0.3 {\footnotesize $\pm$3.0} & $+$67.3 {\footnotesize $\pm$6.8} & $+$67.5 {\footnotesize $\pm$6.1} & $+$67.6 {\footnotesize $\pm$2.8} \\
SmolVLA $-$ SmolVLA-LP & $-$4.2 {\footnotesize $\pm$3.9} & $-$24.1 {\footnotesize $\pm$7.5} & $-$22.4 {\footnotesize $\pm$7.3} & $-$28.7 {\footnotesize $\pm$3.8} \\
SmolVLA $-$ X-VLA & $-$3.5 {\footnotesize $\pm$4.4} & $+$14.8 {\footnotesize $\pm$7.3} & $+$14.2 {\footnotesize $\pm$6.6} & $+$9.7 {\footnotesize $\pm$4.1} \\
SmolVLA-LP $-$ X-VLA & $+$0.7 {\footnotesize $\pm$2.8} & $+$38.9 {\footnotesize $\pm$8.4} & $+$36.6 {\footnotesize $\pm$6.6} & $+$38.4 {\footnotesize $\pm$3.8} \\
\midrule
Pairs ordered as on the held-out placements & 8 & 10 & 10 & \\
Pairs whose interval excludes zero & 5 & 10 & 10 & 10 \\
\bottomrule
\end{tabular}

%% file: tables/app_adaptation.tex
\begin{tabular}{@{}lllcclc@{}}
\toprule
Demonstrations & Trained part & Updates & Standard & Ref.-selected & \multicolumn{1}{c}{Change} & Within \\
\midrule
\multicolumn{7}{l}{\textit{Task 0}} \\
\multicolumn{3}{l}{Base model} & 97.0 & 74.0 & -- &  \\
All 200 & Whole model & Fixed & 100.0 & 29.0 & $-$45.0 {\footnotesize [$-$55.7, $-$34.1]} & 10\,pp \\
All 200 & Action expert & Fixed & 97.5 & 12.1 & $-$61.9 {\footnotesize [$-$70.3, $-$53.2]} & 5\,pp \\
First 50 & Whole model & Fixed & 100.0 & 31.4 & $-$42.6 {\footnotesize [$-$54.0, $-$30.1]} & 10\,pp \\
Random 50 & Whole model & Fixed & 99.0 & 30.0 & $-$44.0 {\footnotesize [$-$57.1, $-$31.0]} & 10\,pp \\
Random 50 & Action expert & Fixed & 97.9 & 11.5 & $-$62.5 {\footnotesize [$-$70.4, $-$53.9]} & 10\,pp \\
Random 50 & Whole model & Scaled & 97.2 & 49.0 & $-$25.0 {\footnotesize [$-$33.6, $-$16.2]} & 10\,pp \\
Random 50 & Action expert & Scaled & 92.3 & 6.8 & $-$67.2 {\footnotesize [$-$74.7, $-$59.1]} & No \\
\midrule
\multicolumn{7}{l}{\textit{Task 5}} \\
\multicolumn{3}{l}{Base model} & 100.0 & 93.3 & -- &  \\
All 200 & Whole model & Fixed & 100.0 & 34.7 & $-$58.6 {\footnotesize [$-$77.8, $-$42.5]} & 5\,pp \\
All 200 & Action expert & Fixed & 99.5 & 16.1 & $-$77.2 {\footnotesize [$-$89.2, $-$63.6]} & 5\,pp \\
First 50 & Whole model & Fixed & 100.0 & 4.2 & $-$89.2 {\footnotesize [$-$96.7, $-$79.2]} & 5\,pp \\
Random 50 & Whole model & Fixed & 100.0 & 31.8 & $-$61.6 {\footnotesize [$-$76.1, $-$48.1]} & 5\,pp \\
Random 50 & Action expert & Fixed & 99.4 & 13.2 & $-$80.1 {\footnotesize [$-$90.8, $-$67.5]} & 5\,pp \\
Random 50 & Whole model & Scaled & 90.1 & 26.0 & $-$67.3 {\footnotesize [$-$83.5, $-$51.4]} & No \\
Random 50 & Action expert & Scaled & 95.5 & 12.1 & $-$81.2 {\footnotesize [$-$90.7, $-$69.2]} & No \\
\midrule
\multicolumn{7}{l}{\textit{Task 8}} \\
\multicolumn{3}{l}{Base model} & 97.5 & 97.1 & -- &  \\
All 200 & Whole model & Fixed & 100.0 & 6.8 & $-$90.3 {\footnotesize [$-$95.8, $-$82.9]} & 10\,pp \\
All 200 & Action expert & Fixed & 76.0 & 11.9 & $-$85.1 {\footnotesize [$-$91.3, $-$78.2]} & No \\
First 50 & Whole model & Fixed & 98.2 & 2.4 & $-$94.7 {\footnotesize [$-$98.8, $-$88.8]} & 5\,pp \\
Random 50 & Whole model & Fixed & 98.9 & 6.7 & $-$90.4 {\footnotesize [$-$96.0, $-$83.6]} & 10\,pp \\
Random 50 & Action expert & Fixed & 74.3 & 11.3 & $-$85.7 {\footnotesize [$-$91.7, $-$78.8]} & No \\
Random 50 & Whole model & Scaled & 99.4 & 23.1 & $-$73.9 {\footnotesize [$-$82.2, $-$65.6]} & 10\,pp \\
Random 50 & Action expert & Scaled & 91.3 & 14.5 & $-$82.6 {\footnotesize [$-$88.5, $-$75.5]} & No \\
\bottomrule
\end{tabular}

%% file: tables/app_human_ft.tex
\begin{tabular}{@{}lrlrl@{}}
\toprule
& \multicolumn{2}{c}{Standard} & \multicolumn{2}{c}{Selected} \\
\cmidrule(lr){2-3}\cmidrule(l){4-5}
Task & \multicolumn{1}{c}{Base} & \multicolumn{1}{c}{Change} & \multicolumn{1}{c}{Base} & \multicolumn{1}{c@{}}{Change} \\
\midrule
\multicolumn{5}{@{}l}{\textit{$\pi_{0.5}$}} \\
0 & 97.0 & $-$0.5 {\footnotesize [$-$3.7, $+$2.7]} & 74.0 & $-$7.1 {\footnotesize [$-$14.3, $+$0.0]} \\
1 & 99.5 & $-$3.3 {\footnotesize [$-$10.8, $+$1.0]} & 74.4 & $+$9.1 {\footnotesize [$+$2.4, $+$15.4]} \\
2 & 95.5 & $+$4.2 {\footnotesize [$-$0.3, $+$11.0]} & 80.8 & $-$4.3 {\footnotesize [$-$13.1, $+$4.7]} \\
3 & 98.0 & $-$1.8 {\footnotesize [$-$6.2, $+$1.2]} & 86.7 & $-$12.5 {\footnotesize [$-$21.7, $-$2.8]} \\
4 & 97.5 & $-$2.0 {\footnotesize [$-$9.5, $+$2.2]} & 77.5 & $+$7.5 {\footnotesize [$-$4.2, $+$18.9]} \\
5 & 100.0 & $-$2.0 {\footnotesize [$-$6.2, $+$0.0]} & 93.3 & $-$6.4 {\footnotesize [$-$15.6, $+$1.4]} \\
6 & 100.0 & $-$4.0 {\footnotesize [$-$8.3, $-$1.0]} & 90.8 & $-$9.4 {\footnotesize [$-$21.1, $+$1.7]} \\
7 & 99.5 & $-$0.5 {\footnotesize [$-$2.5, $+$1.2]} & 95.0 & $-$21.1 {\footnotesize [$-$39.4, $-$6.7]} \\
8 & 97.5 & $+$1.3 {\footnotesize [$-$1.2, $+$4.7]} & 97.1 & $-$14.9 {\footnotesize [$-$21.0, $-$9.0]} \\
9 & 100.0 & $-$0.7 {\footnotesize [$-$2.0, $+$0.0]} & 85.0 & $-$6.9 {\footnotesize [$-$18.6, $+$3.3]} \\
Mean, $h=5$ &  & $-$0.9 {\footnotesize [$-$2.3, $+$0.4]} &  & $-$6.6 {\footnotesize [$-$10.2, $-$3.5]} \\
Mean, $h=10$ &  & $-$0.1 {\footnotesize [$-$1.3, $+$0.8]} &  & $-$4.1 {\footnotesize [$-$6.8, $-$1.2]} \\
\midrule
\multicolumn{5}{@{}l}{\textit{SmolVLA}} \\
0 & 80.5 & $+$3.3 {\footnotesize [$-$6.0, $+$12.3]} & 19.7 & $-$3.8 {\footnotesize [$-$8.7, $+$1.3]} \\
1 & 93.0 & $+$3.7 {\footnotesize [$+$0.0, $+$8.2]} & 40.6 & $+$3.0 {\footnotesize [$-$5.6, $+$11.3]} \\
2 & 91.0 & $+$3.7 {\footnotesize [$-$3.5, $+$12.0]} & 15.8 & $-$2.5 {\footnotesize [$-$9.2, $+$4.6]} \\
3 & 93.0 & $-$0.5 {\footnotesize [$-$9.0, $+$6.3]} & 20.8 & $+$1.7 {\footnotesize [$-$7.8, $+$10.6]} \\
4 & 90.5 & $+$7.3 {\footnotesize [$-$1.0, $+$18.5]} & 31.7 & $+$1.7 {\footnotesize [$-$8.6, $+$12.2]} \\
5 & 75.0 & $+$8.8 {\footnotesize [$-$0.2, $+$19.0]} & 22.5 & $+$5.6 {\footnotesize [$-$4.7, $+$16.7]} \\
6 & 92.5 & $-$2.0 {\footnotesize [$-$9.0, $+$3.5]} & 65.8 & $-$3.1 {\footnotesize [$-$15.3, $+$8.3]} \\
7 & 95.0 & $+$3.8 {\footnotesize [$-$2.5, $+$14.8]} & 40.0 & $-$8.3 {\footnotesize [$-$25.0, $+$8.3]} \\
8 & 82.5 & $+$3.3 {\footnotesize [$-$2.8, $+$10.8]} & 46.3 & $+$2.2 {\footnotesize [$-$3.3, $+$7.4]} \\
9 & 96.5 & $-$3.3 {\footnotesize [$-$10.5, $+$2.3]} & 40.0 & $-$5.8 {\footnotesize [$-$19.2, $+$7.5]} \\
Mean, $h=5$ &  & $+$2.8 {\footnotesize [$+$0.2, $+$5.3]} &  & $-$0.9 {\footnotesize [$-$4.4, $+$2.2]} \\
Mean, $h=10$ &  & $+$1.3 {\footnotesize [$-$1.0, $+$3.5]} &  & $-$2.8 {\footnotesize [$-$6.1, $+$0.2]} \\
\bottomrule
\end{tabular}

%% file: tables/app_smolvla_scripted.tex
\begin{tabular}{@{}lcrlrl@{}}
\toprule
& & \multicolumn{2}{c}{Standard} & \multicolumn{2}{c}{Selected} \\
\cmidrule(lr){3-4}\cmidrule(l){5-6}
$h$ & Task & \multicolumn{1}{c}{Base} & \multicolumn{1}{c}{Change} & \multicolumn{1}{c}{Base} & \multicolumn{1}{c@{}}{Change} \\
\midrule
\multicolumn{6}{@{}l}{\textit{Human demonstrations}} \\
5 & 0 & 80.5 & $+$3.3 {\footnotesize [$-$6.0, $+$12.3]} & 19.7 & $-$3.8 {\footnotesize [$-$8.7, $+$1.3]} \\
5 & 5 & 75.0 & $+$8.8 {\footnotesize [$-$0.2, $+$19.0]} & 22.5 & $+$5.6 {\footnotesize [$-$4.7, $+$16.7]} \\
5 & 8 & 82.5 & $+$3.3 {\footnotesize [$-$2.8, $+$10.8]} & 46.3 & $+$2.2 {\footnotesize [$-$3.3, $+$7.4]} \\
10 & 0 & 87.5 & $+$5.3 {\footnotesize [$-$2.5, $+$13.5]} & 21.7 & $-$3.3 {\footnotesize [$-$10.7, $+$2.9]} \\
10 & 5 & 86.0 & $+$0.3 {\footnotesize [$-$9.5, $+$10.0]} & 32.5 & $+$0.8 {\footnotesize [$-$11.4, $+$13.1]} \\
10 & 8 & 91.0 & $-$1.5 {\footnotesize [$-$8.5, $+$4.2]} & 52.1 & $-$4.0 {\footnotesize [$-$8.9, $+$0.7]} \\
\midrule
\multicolumn{6}{@{}l}{\textit{Scripted demonstrations, recomputed statistics}} \\
5 & 0 & 80.5 & $-$74.2 {\footnotesize [$-$88.7, $-$58.5]} & 19.7 & $-$19.7 {\footnotesize [$-$25.0, $-$14.7]} \\
5 & 5 & 75.0 & $-$3.2 {\footnotesize [$-$23.2, $+$16.0]} & 22.5 & $-$22.5 {\footnotesize [$-$32.5, $-$12.5]} \\
5 & 8 & 82.5 & $-$15.8 {\footnotesize [$-$36.7, $+$4.8]} & 46.3 & $-$43.3 {\footnotesize [$-$50.8, $-$35.3]} \\
10 & 0 & 87.5 & $-$82.8 {\footnotesize [$-$92.5, $-$71.5]} & 21.7 & $-$21.7 {\footnotesize [$-$28.3, $-$15.3]} \\
10 & 5 & 86.0 & $-$23.0 {\footnotesize [$-$39.5, $-$6.2]} & 32.5 & $-$32.5 {\footnotesize [$-$45.8, $-$21.7]} \\
10 & 8 & 91.0 & $+$7.7 {\footnotesize [$-$1.2, $+$20.5]} & 52.1 & $-$47.4 {\footnotesize [$-$54.3, $-$39.6]} \\
\bottomrule
\end{tabular}

%% file: tables/app_fixed_norm.tex
\begin{tabular}{@{}lrcrrcr@{}}
\toprule
\multirow{2}{*}{Task} & \multicolumn{3}{c}{Standard} & \multicolumn{3}{c}{Reference-selected} \\
\cmidrule(lr){2-4}\cmidrule(lr){5-7}
& \multicolumn{1}{c}{Base} & Recomputed & \multicolumn{1}{c}{Fixed} & \multicolumn{1}{c}{Base} & Recomputed & \multicolumn{1}{c@{}}{Fixed} \\
\midrule
0 & 97.0 & 100.0 & 100.0 & 74.0 & 29.0 & 27.2 \\
5 & 100.0 & 100.0 & 100.0 & 93.3 & 34.7 & 29.4 \\
8 & 97.5 & 100.0 & 74.3 & 97.1 & 6.8 & 4.9 \\
\bottomrule
\end{tabular}

%% file: tables/app_norm_contrasts.tex
\begin{tabular}{@{}lrrr@{}}
\toprule
Control & \multicolumn{1}{c}{Task 0} & \multicolumn{1}{c}{Task 5} & \multicolumn{1}{c@{}}{Task 8} \\
\midrule
Fine-tuned, fixed statistics & $-$46.8 {\footnotesize [$-$55.3, $-$38.4]} & $-$63.9 {\footnotesize [$-$77.2, $-$50.0]} & $-$92.2 {\footnotesize [$-$98.5, $-$83.3]} \\
\midrule
Input statistics, no update & $+$2.0 {\footnotesize [$-$4.0, $+$8.3]} & $+$5.8 {\footnotesize [$+$0.0, $+$14.2]} & $-$0.4 {\footnotesize [$-$5.0, $+$5.0]} \\
Output statistics, no update & $-$9.7 {\footnotesize [$-$18.7, $+$0.0]} & $-$47.5 {\footnotesize [$-$57.5, $-$37.5]} & $-$43.3 {\footnotesize [$-$55.0, $-$31.3]} \\
Both, no update & $-$17.3 {\footnotesize [$-$26.0, $-$8.7]} & $-$47.5 {\footnotesize [$-$57.5, $-$37.5]} & $-$42.5 {\footnotesize [$-$52.5, $-$31.3]} \\
Rotation set to zero, no update & $+$2.0 {\footnotesize [$-$6.3, $+$9.7]} & $-$69.2 {\footnotesize [$-$80.0, $-$57.5]} & $-$28.8 {\footnotesize [$-$40.8, $-$17.5]} \\
\bottomrule
\end{tabular}

%% file: appendix/c_instruction.tex
\section{Instruction}
\label{app:instruction}
In this section, we describe how LIBERO-Plus builds its instructions and present the per-model results, the text controls and the second evaluation of Section~\ref{sec:instruction}.

\subsection{The appended text and its removal}
\label{app:census}
LIBERO-Plus builds most variant instructions by appending the parameters that generated the variant to the task sentence, such as the camera, the initial state, the lighting, the layout and the noise level. The variants that rewrite the language carry no such text. Table~\ref{tab:census} counts the variants of each kind from their task metadata. The repair removes the appended parameters and leaves every rewritten instruction unchanged.

\textbf{Rewritten-language variants.}\label{app:by-family} Their instruction is the same in both runs, so their change shows how much the two runs differ for reasons other than the text: at most 1.0 point for every model and suite. Appendix~\ref{app:pairedrng} measures this variation over repeated runs.

\begin{table}[!htbp]
\centering\small
\caption{\textbf{LIBERO-Plus variants whose instruction carries appended parameters.}}
\label{tab:census}
\vspace{4pt}%
\input{tables/app_census}
\end{table}

\subsection{Results for all models}
\label{app:instruction-tables}
\label{app:instruction-matrix}
Table~\ref{tab:instruction-intervals} gives the success of every model with the original and the repaired instruction on every suite, with the change and its interval over variants (Appendix~\ref{app:stats}; \InstrCIDraws{} resamples). The eight models other than the SmolVLA pair change by at most \InstrOtherMax{} points, and \InstrCIOtherExclInstance{} of their \InstrCIOtherCells{} model--suite intervals exclude zero.

\textbf{OpenVLA-OFT.} The two OpenVLA-OFT models run with their own evaluation code, which records no outcome when a run aborts, so they have paired outcomes on \OftPairMin{} to \OftPairMax{} of the \OftSuiteMin{} to \OftSuiteMax{} variants of a suite. Counting every missing outcome as a success or as a failure moves an OpenVLA-OFT change by at most 2.6 points. On the variants that both OpenVLA-OFT models complete under both instructions, \OftSubCoverMin{} to \OftSubCoverMax{}\% of each suite, no model's change moves by more than \OftSubShift{} points.

\subsection{Length-matched filler}
\label{app:length-matched}
The repair shortens the instruction, so we test whether a large loss needs the parameter values themselves. We append to the repaired instruction as many neutral words as the repair removed, drawn from the twenty words of Appendix~\ref{app:suffix}, for the two SmolVLA models on the 100 variants per suite and the three seeded repetitions of Appendix~\ref{app:pairedrng} (Table~\ref{tab:length-matched}).

The filler lowers success in all \LenCells{} model--suite pairs, by \LenMin{} to \LenMax{} points, most of what the parameters cost, so the parameter text itself is not needed for a large loss. Replacing the parameters with the filler moves success by at most \ContentMax{} points, and the interval of this change excludes zero in \ContentResolved{} of the \LenCells{} pairs. The word count is matched but the token count is not: the padded instruction has fewer tokens than the original in \TokFewer{} of the \InstrDistinct{} distinct instruction strings among the 10{,}030 variants, and it still reproduces most of the loss. Both SmolVLA models truncate the instruction at \TokLimit{} tokens, which \TokOver{} of the \InstrDistinct{} original instructions exceeds.

\begin{table}[!htbp]
\centering\small
\caption{\textbf{Original, repaired and padded instructions for the SmolVLA models.} The padding cost is repaired success minus padded success, and the replacement effect is padded success minus original success. Intervals are 95\% $t$ intervals over the three seeded repetitions, on differences paired within a repetition.}
\label{tab:length-matched}
\vspace{4pt}%
\input{tables/app_length_matched}
\end{table}

\subsection{Neutral words after the benchmark's instruction}
\label{app:suffix}
Section~\ref{sec:instruction} appends $k$ ordinary English words, which name no object, direction or action, to the instructions of standard LIBERO-Object. They come from one fixed cycle of twenty adverbs: \emph{perhaps, otherwise, certainly, usually, rather, somewhat, eventually, briefly, apparently, indeed, thereby, likewise, moreover, namely, hence, besides, altogether, instead, presumably, meanwhile}. A hash of the instruction sets its start in the cycle, and its suffix of $k$ words continues from there, so a longer suffix extends every shorter one. For example, \emph{pick up the alphabet soup and place it in the basket} becomes \emph{pick up the alphabet soup and place it in the basket instead presumably} at $k=2$. The LIBERO-Goal control of Appendix~\ref{app:goal-instruction} uses the same rule. Table~\ref{tab:suffix} gives success for every length, each on the 500 standard episodes at each model's default horizon, and the run with $k=0$ reproduces the default of Table~\ref{tab:hcurve} exactly. With the longest suffix, the instructions have \SufTokMin{} to \SufTokMax{} tokens, below the length at which the SmolVLA models truncate.

\begin{table}[!htbp]
\centering\small
\caption{\textbf{Success on LIBERO-Object with neutral words appended.}}
\label{tab:suffix}
\vspace{4pt}%
\pendingsized{\input{tables/app_suffix}}
\end{table}

\FloatBarrier
\Needspace{8\baselineskip}%
Two ordinary words drawn at random in place of the fixed cycle give a similar result. The vocabulary has 200 words that name places, nature, culture and time, with adjectives of weather and era and adverbs of time, frequency and stance, and it leaves out every word of the benchmark's instructions and words of manner, speed, size or direction. Each episode appends two distinct words from it, and each of three independent draws covers the 500 standard episodes of every model at its default horizon (Table~\ref{tab:random-words}). Averaged over the draws, SmolVLA loses \RWSmolPooled{} points and SmolVLA-LP \RWSmolPlusPooled{}, comparable to their losses of \RWSmolFiller{} and \RWSmolPlusFiller{} points with two words of the fixed cycle, and each of the other six models changes by at most \RWOthersMax{} points.

{\small The 200 words, in alphabetical order: admittedly, alpine, ancient, anniversary, annual, annually, arctic, arguably, arid, astronomer, autumn, autumnal, ballet, baroque, basically, birthday, botanical, breeze, butterfly, canyon, carnival, castle, cathedral, celestial, century, chapter, city, civic, classical, cloud, cloudy, coastal, comet, concert, coral, cosmic, country, cultural, culturally, decade, desert, distant, dolphin, eagle, eclipse, economic, election, elephant, empire, equatorial, evening, evidently, explorer, falcon, festival, festive, foggy, formerly, frankly, frequently, galaxy, giraffe, glacier, gothic, guitar, harbor, hardly, heroic, historic, historically, holiday, however, humid, imperial, island, journey, kangaroo, kingdom, lagoon, lately, legal, lesson, lightning, lion, literary, lunar, marathon, marine, meadow, medical, medieval, melody, meteor, midnight, misty, modern, monarch, monthly, mountain, museum, music, musical, mythical, national, nautical, nebula, nightly, noble, nonetheless, notably, novel, novelist, occasionally, ocean, oceanic, opera, orchard, orchestra, ordinarily, originally, painter, painting, parade, passport, penguin, piano, pilgrim, planet, poet, poetic, poetry, polar, political, politically, portrait, possibly, postcard, prairie, probably, rainbow, rainy, rare, rarely, recently, reef, republic, rhythm, riddle, river, royal, rural, rustic, sailor, science, scientifically, sculpture, season, seasonal, seemingly, seldom, shepherd, snowy, solar, sometimes, sparrow, story, sudden, summer, sunrise, sunset, supposedly, telescope, temple, theater, thunder, tide, tiger, tomorrow, tonight, tournament, tropical, trumpet, urban, vacation, valley, village, vineyard, vintage, violin, vivid, volcanic, volcano, voyage, weekend, weekly, whale, windy, winter, wintry, zebra.}

\begin{table}[!htbp]
\centering\small
\caption{\textbf{Two random words appended on LIBERO-Object.} Success with the original instruction in percent, and the change in points after appending two words of the fixed cycle or two random words in each of three draws. The mean of the draws has a paired 95\% interval over initial states.}
\label{tab:random-words}
\vspace{4pt}%
\pendingsized{\input{tables/app_random_words}}
\end{table}

\FloatBarrier
\subsection{Instructions on LIBERO-Goal}
\label{app:goal-instruction}
LIBERO-Goal places all ten tasks in one shared scene. We run the 500 standard episodes of each model under five instructions: the task's own, the same with two or eight neutral words appended, the instruction of another task, with task $i$ given that of task $i+1$ modulo ten, and an empty string. Success means completion of the evaluated task, and under another task's instruction we also record whether the named task is completed. A task counts as completed when its goal predicate, false in the initial state, becomes true within the step budget, which is the same for all five instructions. Both predicates are checked after every action of the same episode, and episodes in which a predicate already holds at the start stay in the denominator. Table~\ref{tab:goal-instruction} gives the results. Given another task's instruction, every model completes the named task in \GoalInstructedMin{} to \GoalInstructedMax{}\% of episodes and the evaluated task in none. Two neutral words change the success of SmolVLA and SmolVLA-LP by $\GoalSmolVLASfxTwoDelta$ and $\GoalSmolVLAPlusSfxTwoDelta$ points, eight words change that of $\piz$ by $\GoalPiZeroSfxEightDelta$ points and that of the other five models by at most \GoalOtherEightMax{} points. With an empty instruction, every model succeeds in fewer than one in ten episodes.

\begin{table}[!htbp]
\centering\small
\caption{\textbf{Success on LIBERO-Goal under five instructions.} ``Named task'' is the percentage of episodes in which the task named in the given instruction is completed.}
\label{tab:goal-instruction}
\vspace{4pt}%
\input{tables/app_goal}
\end{table}

\FloatBarrier
\subsection{Replication with seeded episodes}
\label{app:pairedrng}
Section~\ref{sec:instruction} reports that the large instruction effects replicate in a second evaluation, which reruns $\piz$, $\pif$, SmolVLA and SmolVLA-LP on 100 variants per suite under both instructions, at 50 and at 10 executed actions. The variants are taken in equal numbers from each task label of the suite, at evenly spaced positions in the benchmark's listing: ten from each of the ten labels, and on Goal, where the severity variants carry ten labels of their own, five from each of the twenty. Each of three repetitions has its own seed, and at every reset the random number generators are reseeded from the repetition's seed, the variant and the episode, so the two instructions and the two horizons of a variant are paired within a repetition. Intervals are 95\% $t$-intervals over the three repetitions. Table~\ref{tab:pairedrng} gives the change at each horizon and the difference between the two.

We count an effect of more than 5 points in the full study as reproduced when the change at 50 actions has the same sign and its interval excludes zero. All \PRCriterionTested{} such effects are reproduced. No difference between the two horizons has an interval that excludes zero. On Goal, half of the subset consists of severity variants (labeled with a perturbation level from 1 to 5), which make up \GoalSeverityPop{}\% of the suite.

\begin{table}[!htbp]
\centering\small
\caption{\textbf{Instruction repair in the second evaluation.} An asterisk marks an effect of more than 5 points in the full study.}
\label{tab:pairedrng}
\label{tab:pairedrng-interaction}
\vspace{4pt}%
\input{tables/app_second_eval}
\end{table}
\FloatBarrier

The replication also runs every model at its default horizon, including VLA-JEPA, MolmoAct2, X-VLA, EVO-1 and the two OpenVLA-OFT models (Table~\ref{tab:seeded-all}), with a bootstrap over the variants of each suite that keeps the three seeds of a variant together. Pooled over the four suites, removing the appended text raises SmolVLA by \PRSmolAll{} points and SmolVLA-LP by \PRSmolPlusAll{} points. On any single suite, the repair changes every other model by at most \PROthersSuiteMax{} points, and the pooled effect of each model varies across the three seeds with a standard deviation of at most \PRSeedSDMax{} points.

\begin{table}[!htbp]
\centering\small
\caption{\textbf{Instruction repair in the second evaluation, all models.} Every model runs at its default horizon. The change is repaired minus original success, in points, per suite and pooled over the four suites with a 95\% interval, and the last column is the standard deviation of the pooled change across the three seeds.}
\label{tab:seeded-all}
\vspace{4pt}%
\pendingsized{\input{tables/app_seeded_all}}
\end{table}

\begin{table}[!htbp]
\centering\small
\caption{\textbf{Success with the original and the repaired instruction on LIBERO-Plus.}}
\label{tab:instruction-intervals}
\label{tab:instruction-full}
\label{tab:oft-coverage}
\vspace{4pt}%
\input{tables/app_instruction_all}
\end{table}
\FloatBarrier

%% file: tables/app_census.tex
\begin{tabular}{@{}lrcc@{}}
\toprule
Suite & \multicolumn{1}{c}{Variants} & With appended text & Rewritten language \\
\midrule
Object & 2,518 & 2,164 & 354 \\
Spatial & 2,402 & 2,012 & 390 \\
Goal & 2,591 & 2,181 & 410 \\
Long & 2,519 & 2,136 & 383 \\
Total & 10,030 & 8,493 & 1,537 \\
\bottomrule
\end{tabular}

%% file: tables/app_length_matched.tex
\begin{tabular}{@{}llccrll@{}}
\toprule
Model & Suite & Original & Repaired & \multicolumn{1}{c}{Padded} & \multicolumn{1}{c}{Padding cost} & \multicolumn{1}{c@{}}{Replacement} \\
\midrule
\multirow{4}{*}{SmolVLA} & Object & 14.3 & 31.0 & 13.3 & +17.7 {\footnotesize [+14.8, +20.5]} & $-$1.0 {\footnotesize [$-$3.5, +1.5]} \\
 & Spatial & 17.0 & 39.0 & 13.3 & +25.7 {\footnotesize [+15.3, +36.0]} & $-$3.7 {\footnotesize [$-$9.4, +2.1]} \\
 & Goal & 1.7 & 34.0 & 5.0 & +29.0 {\footnotesize [+26.5, +31.5]} & +3.3 {\footnotesize [+1.9, +4.8]} \\
 & Long & 9.3 & 24.0 & 12.3 & +11.7 {\footnotesize [+4.1, +19.3]} & +3.0 {\footnotesize [$-$4.5, +10.5]} \\
\addlinespace
\multirow{4}{*}{SmolVLA-LP} & Object & 10.0 & 62.3 & 11.3 & +51.0 {\footnotesize [+46.7, +55.3]} & +1.3 {\footnotesize [$-$4.4, +7.1]} \\
 & Spatial & 18.0 & 46.7 & 19.3 & +27.3 {\footnotesize [+17.3, +37.4]} & +1.3 {\footnotesize [$-$12.8, +15.5]} \\
 & Goal & 7.7 & 34.3 & 7.7 & +26.7 {\footnotesize [+22.9, +30.5]} & +0.0 {\footnotesize [$-$4.3, +4.3]} \\
 & Long & 13.3 & 51.3 & 16.3 & +35.0 {\footnotesize [+28.4, +41.6]} & +3.0 {\footnotesize [+0.5, +5.5]} \\
\bottomrule
\end{tabular}

%% file: tables/app_suffix.tex
\begin{tabular}{@{}lrrrr@{}}
\toprule
\multirow{2}{*}{Model} & \multicolumn{4}{c}{Words appended} \\
\cmidrule(lr){2-5}
 & \multicolumn{1}{c}{0} & \multicolumn{1}{c}{2} & \multicolumn{1}{c}{8} & \multicolumn{1}{c@{}}{32} \\
\midrule
$\pi_0$ & 65.4 & 60.6 & 64.8 & 59.6 \\
$\pi_{0.5}$ & 83.2 & 82.6 & 81.0 & 66.8 \\
SmolVLA & 69.0 & 37.2 & 4.0 & 0.4 \\
SmolVLA-LP & 87.4 & 11.4 & 4.2 & 0.0 \\
VLA-JEPA & 99.8 & 99.8 & 100.0 & 100.0 \\
MolmoAct2 & 99.8 & 99.8 & 99.0 & 99.0 \\
EVO-1 & 83.6 & 83.8 & 85.0 & 84.6 \\
X-VLA & 99.2 & 99.2 & 99.4 & 99.2 \\
\bottomrule
\end{tabular}

%% file: tables/app_random_words.tex
\begin{tabular}{@{}lccrrrl@{}}
\toprule
& Original & Two neutral & \multicolumn{4}{c}{Two random words} \\
\cmidrule(l){4-7}
Model & instruction & words & \multicolumn{1}{c}{Draw 1} & \multicolumn{1}{c}{Draw 2} & \multicolumn{1}{c}{Draw 3} & \multicolumn{1}{c@{}}{Mean of draws} \\
\midrule
$\pi_0$ & 65.4 & $-$4.8 & $-$3.8 & $-$4.0 & $-$5.8 & $-$4.5 {\footnotesize [$-$8.8, $-$0.3]} \\
$\pi_{0.5}$ & 83.2 & $-$0.6 & $+$0.8 & $-$0.8 & $-$1.4 & $-$0.5 {\footnotesize [$-$3.3, $+$2.5]} \\
SmolVLA & 69.0 & $-$31.8 & $-$37.2 & $-$34.8 & $-$37.0 & $-$36.3 {\footnotesize [$-$41.0, $-$31.6]} \\
SmolVLA-LP & 87.4 & $-$76.0 & $-$68.4 & $-$68.6 & $-$70.2 & $-$69.1 {\footnotesize [$-$72.3, $-$65.6]} \\
X-VLA & 99.2 & $+$0.0 & $+$0.6 & $+$0.2 & $+$0.4 & $+$0.4 {\footnotesize [$-$0.3, $+$1.3]} \\
VLA-JEPA & 99.8 & $+$0.0 & $+$0.2 & $+$0.2 & $+$0.2 & $+$0.2 {\footnotesize [$+$0.0, $+$0.6]} \\
MolmoAct2 & 99.8 & $+$0.0 & $+$0.0 & $-$0.4 & $+$0.2 & $-$0.1 {\footnotesize [$-$0.5, $+$0.5]} \\
EVO-1 & 83.6 & $+$0.2 & $-$2.0 & $-$2.0 & $+$1.0 & $-$1.0 {\footnotesize [$-$4.1, $+$2.1]} \\
\bottomrule
\end{tabular}

%% file: tables/app_goal.tex
\begin{tabular}{@{}lrrrccc@{}}
\toprule
\multirow{2}{*}{Model} & \multicolumn{1}{c}{\multirow{2}{*}{Own}} & \multicolumn{2}{c}{Neutral words} & \multicolumn{2}{c}{Another task's instruction} & \multirow{2}{*}{Empty} \\
\cmidrule(lr){3-4}\cmidrule(lr){5-6}
 & & \multicolumn{1}{c}{2} & \multicolumn{1}{c}{8} & Own task & Named task & \\
\midrule
$\pi_0$ & 81.2 & 78.6 & 68.8 & 0.0 & 75.8 & 0.0 \\
$\pi_{0.5}$ & 87.4 & 87.8 & 84.2 & 0.0 & 86.4 & 4.8 \\
SmolVLA & 77.6 & 16.0 & 2.4 & 0.0 & 82.8 & 0.0 \\
SmolVLA-LP & 87.2 & 36.0 & 1.8 & 0.0 & 86.0 & 0.0 \\
VLA-JEPA & 96.8 & 97.2 & 97.6 & 0.0 & 97.4 & 4.6 \\
MolmoAct2 & 97.6 & 97.8 & 95.6 & 0.0 & 97.8 & 5.8 \\
EVO-1 & 89.0 & 84.4 & 81.8 & 0.0 & 89.6 & 3.6 \\
X-VLA & 97.6 & 98.8 & 98.2 & 0.0 & 98.8 & 9.2 \\
\bottomrule
\end{tabular}

%% file: tables/app_second_eval.tex
\begin{tabular}{@{}lllll@{}}
\toprule
Model & Suite & \multicolumn{1}{c}{Change, 50 actions} & \multicolumn{1}{c}{Change, 10 actions} & \multicolumn{1}{c@{}}{Difference} \\
\midrule
\multirow{4}{*}{$\pi_0$} & Object & +3.7 {\footnotesize [$-$2.6, +9.9]} & +8.0 {\footnotesize [+3.7, +12.3]} & +4.3 {\footnotesize [$-$2.8, +11.5]} \\
 & Spatial & +3.7 {\footnotesize [+0.8, +6.5]} & $-$0.3 {\footnotesize [$-$5.5, +4.8]} & $-$4.0 {\footnotesize [$-$11.5, +3.5]} \\
 & Goal & +8.0 {\footnotesize [+3.0, +13.0]}$^\ast$ & +6.7 {\footnotesize [+0.9, +12.4]} & $-$1.3 {\footnotesize [$-$8.9, +6.3]} \\
 & Long & +5.3 {\footnotesize [$-$3.4, +14.1]} & +7.7 {\footnotesize [$-$4.6, +19.9]} & +2.3 {\footnotesize [$-$12.8, +17.5]} \\
\addlinespace
\multirow{4}{*}{$\pi_{0.5}$} & Object & +1.3 {\footnotesize [$-$10.1, +12.8]} & $-$1.7 {\footnotesize [$-$3.1, $-$0.2]} & $-$3.0 {\footnotesize [$-$13.8, +7.8]} \\
 & Spatial & $-$1.0 {\footnotesize [$-$5.3, +3.3]} & $-$1.0 {\footnotesize [$-$5.3, +3.3]} & +0.0 {\footnotesize [$-$4.3, +4.3]} \\
 & Goal & +2.0 {\footnotesize [$-$7.0, +11.0]} & +0.0 {\footnotesize [$-$4.3, +4.3]} & $-$2.0 {\footnotesize [$-$12.8, +8.8]} \\
 & Long & +6.3 {\footnotesize [$-$5.9, +18.6]} & +0.0 {\footnotesize [$-$4.3, +4.3]} & $-$6.3 {\footnotesize [$-$14.3, +1.7]} \\
\addlinespace
\multirow{4}{*}{SmolVLA} & Object & +16.7 {\footnotesize [+15.2, +18.1]}$^\ast$ & +18.3 {\footnotesize [+5.8, +30.8]} & +1.7 {\footnotesize [$-$9.5, +12.9]} \\
 & Spatial & +22.0 {\footnotesize [+7.1, +36.9]}$^\ast$ & +12.3 {\footnotesize [$-$3.4, +28.1]} & $-$9.7 {\footnotesize [$-$39.3, +20.0]} \\
 & Goal & +32.3 {\footnotesize [+30.9, +33.8]}$^\ast$ & +32.3 {\footnotesize [+28.5, +36.1]} & +0.0 {\footnotesize [$-$4.3, +4.3]} \\
 & Long & +14.7 {\footnotesize [+13.2, +16.1]}$^\ast$ & +20.3 {\footnotesize [+12.3, +28.3]} & +5.7 {\footnotesize [$-$1.5, +12.8]} \\
\addlinespace
\multirow{4}{*}{SmolVLA-LP} & Object & +52.3 {\footnotesize [+47.2, +57.5]}$^\ast$ & +49.3 {\footnotesize [+47.9, +50.8]} & $-$3.0 {\footnotesize [$-$8.0, +2.0]} \\
 & Spatial & +28.7 {\footnotesize [+9.4, +48.0]}$^\ast$ & +19.3 {\footnotesize [+16.5, +22.2]} & $-$9.3 {\footnotesize [$-$29.6, +10.9]} \\
 & Goal & +26.7 {\footnotesize [+25.2, +28.1]}$^\ast$ & +31.3 {\footnotesize [+25.1, +37.6]} & +4.7 {\footnotesize [$-$1.6, +10.9]} \\
 & Long & +38.0 {\footnotesize [+33.7, +42.3]}$^\ast$ & +42.7 {\footnotesize [+23.4, +62.0]} & +4.7 {\footnotesize [$-$16.2, +25.5]} \\
\bottomrule
\end{tabular}

%% file: tables/app_seeded_all.tex
\begin{tabular}{@{}lrrrrlc@{}}
\toprule
& \multicolumn{4}{c}{Change per suite} & & SD across \\
\cmidrule(lr){2-5}
Model & \multicolumn{1}{c}{Object} & \multicolumn{1}{c}{Spatial} & \multicolumn{1}{c}{Goal} & \multicolumn{1}{c}{Long} & \multicolumn{1}{c}{Four suites} & seeds \\
\midrule
$\pi_0$ & $+$3.7 & $+$3.7 & $+$8.0 & $+$5.3 & $+$5.2 {\footnotesize [$+$2.8, $+$7.6]} & 1.5 \\
$\pi_{0.5}$ & $+$1.3 & $-$1.0 & $+$2.0 & $+$6.3 & $+$2.2 {\footnotesize [$+$0.3, $+$4.2]} & 1.6 \\
SmolVLA & $+$16.7 & $+$22.0 & $+$32.3 & $+$14.7 & $+$21.4 {\footnotesize [$+$17.1, $+$25.8]} & 1.5 \\
SmolVLA-LP & $+$52.3 & $+$28.7 & $+$26.7 & $+$38.0 & $+$36.4 {\footnotesize [$+$32.0, $+$40.9]} & 2.3 \\
X-VLA & $+$0.3 & $+$0.7 & $+$2.0 & $-$0.7 & $+$0.6 {\footnotesize [$-$0.8, $+$2.0]} & 1.4 \\
VLA-JEPA & $+$4.0 & $-$1.7 & $+$2.0 & $-$0.7 & $+$0.9 {\footnotesize [$-$1.2, $+$2.9]} & 0.9 \\
MolmoAct2 & $+$1.7 & $+$1.0 & $+$8.7 & $-$1.3 & $+$2.5 {\footnotesize [$+$0.3, $+$4.7]} & 1.4 \\
EVO-1 & $+$5.7 & $-$2.7 & $+$1.3 & $-$0.3 & $+$1.0 {\footnotesize [$-$1.1, $+$3.1]} & 1.6 \\
OpenVLA-OFT & $-$0.3 & $+$1.7 & $+$2.0 & $+$7.0 & $+$2.6 {\footnotesize [$-$0.3, $+$5.5]} & 0.3 \\
OpenVLA-OFT-LP & $-$2.0 & $+$2.3 & $-$1.3 & $+$1.3 & $+$0.1 {\footnotesize [$-$1.9, $+$2.2]} & 0.1 \\
\bottomrule
\end{tabular}

%% file: tables/app_instruction_all.tex
\begin{tabular}{@{}llcclc@{}}
\toprule
Model & Suite & Original & Repaired & \multicolumn{1}{c}{Change} & Paired variants \\
\midrule
\multirow{4}{*}{$\pi_0$} & Object & 43.9 & 46.7 & +2.8 {\footnotesize [+0.9, +4.6]} & 2,518 \\
 & Spatial & 49.0 & 49.6 & +0.6 {\footnotesize [$-$1.5, +2.8]} & 2,402 \\
 & Goal & 48.1 & 56.3 & +8.2 {\footnotesize [+6.2, +10.2]} & 2,591 \\
 & Long & 27.1 & 29.7 & +2.6 {\footnotesize [+0.5, +4.7]} & 2,519 \\
\addlinespace
\multirow{4}{*}{$\pi_{0.5}$} & Object & 54.2 & 55.3 & +1.1 {\footnotesize [$-$0.5, +2.7]} & 2,518 \\
 & Spatial & 66.9 & 66.7 & $-$0.2 {\footnotesize [$-$1.8, +1.2]} & 2,402 \\
 & Goal & 64.2 & 65.0 & +0.8 {\footnotesize [$-$0.7, +2.4]} & 2,591 \\
 & Long & 65.3 & 64.7 & $-$0.6 {\footnotesize [$-$2.1, +0.9]} & 2,519 \\
\addlinespace
\multirow{4}{*}{SmolVLA} & Object & 14.9 & 37.2 & +22.3 {\footnotesize [+20.5, +24.1]} & 2,518 \\
 & Spatial & 17.7 & 35.4 & +17.7 {\footnotesize [+15.5, +19.9]} & 2,402 \\
 & Goal & 3.8 & 44.7 & +40.8 {\footnotesize [+38.9, +42.8]} & 2,591 \\
 & Long & 11.0 & 29.2 & +18.3 {\footnotesize [+16.4, +20.0]} & 2,519 \\
\addlinespace
\multirow{4}{*}{SmolVLA-LP} & Object & 7.3 & 59.0 & +51.7 {\footnotesize [+49.7, +53.7]} & 2,518 \\
 & Spatial & 21.8 & 50.2 & +28.4 {\footnotesize [+26.0, +30.8]} & 2,402 \\
 & Goal & 8.0 & 56.6 & +48.6 {\footnotesize [+46.6, +50.6]} & 2,591 \\
 & Long & 13.0 & 51.2 & +38.2 {\footnotesize [+35.9, +40.3]} & 2,519 \\
\addlinespace
\multirow{4}{*}{VLA-JEPA} & Object & 80.4 & 83.0 & +2.5 {\footnotesize [+1.5, +3.5]} & 2,518 \\
 & Spatial & 87.7 & 86.8 & $-$0.8 {\footnotesize [$-$2.0, +0.3]} & 2,402 \\
 & Goal & 76.3 & 78.4 & +2.1 {\footnotesize [+1.0, +3.2]} & 2,591 \\
 & Long & 66.5 & 67.6 & +1.2 {\footnotesize [$-$0.2, +2.5]} & 2,519 \\
\addlinespace
\multirow{4}{*}{MolmoAct2} & Object & 87.4 & 89.8 & +2.4 {\footnotesize [+1.4, +3.5]} & 2,518 \\
 & Spatial & 91.6 & 91.0 & $-$0.6 {\footnotesize [$-$1.5, +0.4]} & 2,402 \\
 & Goal & 80.0 & 81.8 & +1.8 {\footnotesize [+0.7, +2.9]} & 2,591 \\
 & Long & 80.3 & 81.6 & +1.3 {\footnotesize [$-$0.1, +2.7]} & 2,519 \\
\addlinespace
\multirow{4}{*}{EVO-1} & Object & 42.2 & 45.7 & +3.5 {\footnotesize [+2.2, +4.8]} & 2,518 \\
 & Spatial & 51.9 & 51.5 & $-$0.3 {\footnotesize [$-$1.7, +1.0]} & 2,402 \\
 & Goal & 47.3 & 50.3 & +3.0 {\footnotesize [+1.6, +4.5]} & 2,591 \\
 & Long & 45.4 & 47.1 & +1.7 {\footnotesize [+0.2, +3.3]} & 2,519 \\
\addlinespace
\multirow{4}{*}{X-VLA} & Object & 70.3 & 70.5 & +0.2 {\footnotesize [$-$0.5, +1.0]} & 2,518 \\
 & Spatial & 73.7 & 74.3 & +0.6 {\footnotesize [$-$0.3, +1.5]} & 2,402 \\
 & Goal & 68.3 & 68.8 & +0.5 {\footnotesize [$-$0.6, +1.7]} & 2,591 \\
 & Long & 69.4 & 67.7 & $-$1.7 {\footnotesize [$-$2.8, $-$0.6]} & 2,519 \\
\addlinespace
\multirow{4}{*}{OpenVLA-OFT} & Object & 68.9 & 69.4 & +0.5 {\footnotesize [$-$0.7, +1.6]} & 2,513 \\
 & Spatial & 85.5 & 87.0 & +1.5 {\footnotesize [+0.6, +2.5]} & 2,402 \\
 & Goal & 64.0 & 68.4 & +4.4 {\footnotesize [+2.9, +5.8]} & 2,590 \\
 & Long & 68.0 & 68.5 & +0.5 {\footnotesize [$-$0.9, +1.9]} & 2,423 \\
\addlinespace
\multirow{4}{*}{OpenVLA-OFT-LP} & Object & 83.2 & 83.4 & +0.2 {\footnotesize [$-$0.6, +1.0]} & 2,517 \\
 & Spatial & 86.4 & 87.4 & +1.0 {\footnotesize [+0.2, +1.8]} & 2,396 \\
 & Goal & 70.0 & 70.0 & +0.0 {\footnotesize [$-$1.0, +1.0]} & 2,587 \\
 & Long & 82.5 & 82.5 & +0.0 {\footnotesize [$-$1.2, +1.2]} & 2,482 \\
\bottomrule
\end{tabular}

%% file: appendix/d_execution.tex
\section{Executed Actions}
\label{app:horizon}
This section gives the full execution grid of Section~\ref{sec:execution} on standard LIBERO and in SimplerEnv, and Appendix~\ref{app:external} compares SimplerEnv orderings with real-robot rankings.

\textbf{Standard LIBERO.} Table~\ref{tab:hcurve} gives the success of every model at 5, 10 and 25 executed actions and at its default. We use ten as the common setting because every model in the study supports it and it lies within the range used by the models' own evaluations, 5 to 30 actions. The public $\piz$ and $\pif$ configurations execute 50 actions, while the LIBERO example in the models' own code base executes five~\citep{openpi2025}. Success under the two settings differs by \RouteFiveMin{} to \RouteFiveMax{} points on $\pif$ and by \RouteZeroMin{} to \RouteZeroMax{} points on $\piz$ across the four suites.

\textbf{Published results.} The LeRobot LIBERO documentation\footnote{\url{https://huggingface.co/docs/lerobot/libero}, section ``Reproducing published results''.} reports results of $\pif$ at ten executed actions for \texttt{lerobot/pi05\_libero\_finetuned}, obtained with LeRobot's evaluation loop. In the LeRobot version we use (0.5.2), this loop skips an initial state after each success, so the states it evaluates depend on the outcomes. With that loop and the checkpoint of Table~\ref{tab:checkpoints}, $\pif$ reaches \ParityLoopObject{}, \ParityLoopSpatial{}, \ParityLoopGoal{} and \ParityLoopLong{} on Object, Spatial, Goal and Long, against \ParityDocObject{}, \ParityDocSpatial{}, \ParityDocGoal{} and \ParityDocLong{} in the documentation, and Table~\ref{tab:hcurve} gives the same setting on the fixed initial states of Appendix~\ref{app:protocol}. With its own evaluation code, OpenVLA-OFT reaches 97.0, 99.0, 97.6 and 94.8, against the reported 98.4, 97.6, 97.9 and 94.5.

\begin{table}[!htbp]
\centering\small
\caption{\textbf{Success on standard LIBERO at each number of executed actions.} The default is 50 for every model except X-VLA, whose default is 30. {At LIBERO's 20\,Hz, $h$ actions last $h/20$\,s of simulated time.}}
\label{tab:hcurve}
\label{tab:e0}
\vspace{4pt}%
\pendingsized{\input{tables/app_horizon_libero}}
\end{table}
\subsection{Execution responses in SimplerEnv}
\label{app:simplerenv}
The SimplerEnv evaluation code~\citep{li2024simpler} sets the number of actions a model executes between observations, so the intervention of Section~\ref{sec:execution} applies without change. We evaluate the X-VLA checkpoints for WidowX and for Google Robot~\citep{zheng2025xvla} and DB-CogACT on WidowX~\citep{li2024cogact,xie2025dexbotic}, which predicts 16 actions per call. Table~\ref{tab:simplerenv-horizon} gives the mean success over each configuration's tasks. Each of the three configurations performs worst at its shortest setting, and one of them peaks in the middle of its range. \label{app:reproduction}As a reproduction check, a separate run of X-VLA with the distributed evaluation code succeeds in 92 of 96 episodes on the WidowX visual-matching set, against 91 of 96 in the outcomes distributed with that code.

\begin{table}[!htbp]
\centering\small
\caption{\textbf{Success on SimplerEnv at each number of executed actions.} {WidowX runs at 5\,Hz and Google Robot at 3\,Hz, so $h$ actions last $h/5$ and $h/3$\,s of simulated time.}}
\label{tab:simplerenv-horizon}
\vspace{4pt}%
\input{tables/app_simplerenv}
\end{table}

\subsection{Predictive validity for real-robot rankings}
\label{app:external}
A separate study, with its own checkpoints, asks whether success on reference-selected conditions predicts real-robot rankings. For five checkpoints over 3{,}876 episodes, it compares how well success on reference-selected translated conditions in SimplerEnv and the standard score predict published real-robot orderings. Our SimplerEnv runs reproduce the published simulator results, and the reference succeeds on the kept conditions. Against the real-robot success rates, both simulator orderings have a Kendall's $\tau$ of \RealTauStd{}. The mean maximum rank violation~\citep{li2024simpler} is \RealMMRVStd{} for the standard score and \RealMMRVRef{} on the reference-selected conditions, and the 95\% interval of the difference, from Jeffreys posteriors of the real-robot success rates with the simulator scores fixed, is [\RealMMRVDiffLo{}, \RealMMRVDiffHi{}].

\FloatBarrier

%% file: tables/app_horizon_libero.tex
\begin{tabular}{@{}lrrrc@{}}
\toprule
\multirow{2}{*}{Model} & \multicolumn{4}{c}{Executed actions $h$} \\
\cmidrule(lr){2-5}
& \multicolumn{1}{c}{$h=5$} & \multicolumn{1}{c}{$h=10$} & \multicolumn{1}{c}{$h=25$} & Default \\
\midrule
\multicolumn{5}{l}{\textit{Object}} \\
\midrule
$\pi_0$ & 75.6 & 80.8 & 79.2 & 65.4 \\
$\pi_{0.5}$ & 97.6 & 99.4 & 97.6 & 83.2 \\
SmolVLA & 90.0 & 92.8 & 84.8 & 69.0 \\
SmolVLA-LP & 96.8 & 98.0 & 94.8 & 87.4 \\
EVO-1 & 93.6 & 99.8 & 96.6 & 83.6 \\
X-VLA & 94.6 & 98.2 & 99.4 & 99.2 \\
\midrule
\multicolumn{5}{l}{\textit{Spatial}} \\
\midrule
$\pi_0$ & 71.6 & 69.0 & 69.8 & 58.8 \\
$\pi_{0.5}$ & 96.8 & 95.4 & 91.0 & 84.0 \\
SmolVLA & 76.6 & 75.4 & 70.8 & 58.8 \\
SmolVLA-LP & 92.4 & 91.6 & 90.2 & 83.4 \\
EVO-1 & 92.2 & 91.6 & 85.2 & 78.6 \\
X-VLA & 87.2 & 94.4 & 92.4 & 92.2 \\
\midrule
\multicolumn{5}{l}{\textit{Goal}} \\
\midrule
$\pi_0$ & 81.8 & 84.6 & 83.4 & 81.2 \\
$\pi_{0.5}$ & 94.6 & 95.2 & 91.0 & 87.4 \\
SmolVLA & 85.6 & 89.0 & 83.6 & 77.6 \\
SmolVLA-LP & 95.0 & 95.4 & 91.8 & 87.2 \\
EVO-1 & 89.8 & 95.4 & 93.6 & 89.0 \\
X-VLA & 86.2 & 93.2 & 97.8 & 97.6 \\
\midrule
\multicolumn{5}{l}{\textit{Long}} \\
\midrule
$\pi_0$ & 42.6 & 43.4 & 38.8 & 38.6 \\
$\pi_{0.5}$ & 94.2 & 94.2 & 93.4 & 82.0 \\
SmolVLA & 65.0 & 70.6 & 64.8 & 46.8 \\
SmolVLA-LP & 82.8 & 85.8 & 82.8 & 66.2 \\
EVO-1 & 86.8 & 91.8 & 89.0 & 76.4 \\
X-VLA & 58.2 & 91.0 & 93.6 & 91.0 \\
\bottomrule
\end{tabular}

%% file: tables/app_simplerenv.tex
\begin{tabular}{@{}llrrrr@{}}
\toprule
\multirow{2}{*}{Model} & \multirow{2}{*}{Robot} & \multicolumn{4}{c}{Executed actions $h$} \\
\cmidrule(lr){3-6}
 & & \multicolumn{1}{c}{1} & \multicolumn{1}{c}{5} & \multicolumn{1}{c}{10} & \multicolumn{1}{c@{}}{30} \\
\midrule
DB-CogACT & WidowX & 8.3 & 42.7 & 45.8 & -- \\
X-VLA & Google Robot & -- & 44.4 & 51.4 & 48.6 \\
X-VLA & WidowX & -- & 0.0 & 30.2 & 93.8 \\
\bottomrule
\end{tabular}

%% file: appendix/e_comparisons.tex
\section{Model Comparisons}
\label{app:comparisons}
This section gives the margins behind the reversals of Section~\ref{sec:comparison}, their resampling by base task, the state comparisons at the default horizon, and the residuals of the common offset.

\subsection{Reversals under the instruction repair and the execution change}
\label{app:ordering-counts}
\begin{figure}[!htbp]
\centering
\pendinggraphics[width=0.8\linewidth]{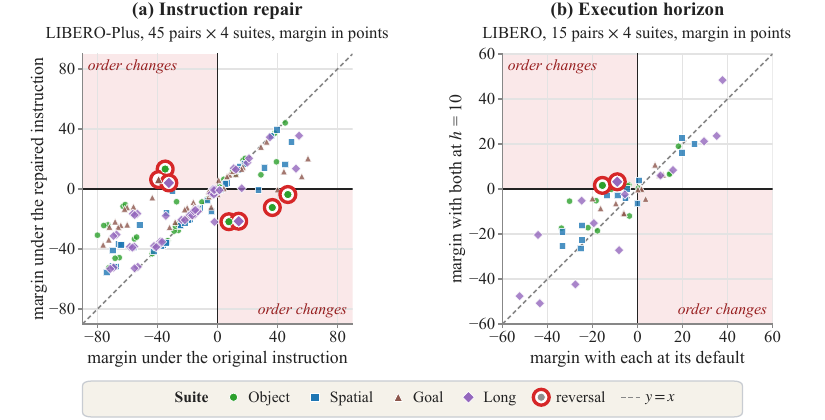}
\caption{\textbf{Margins between models before and after each change.} (a) The instruction repair on LIBERO-Plus. (b) Changing each model's default to ten executed actions on standard LIBERO.}
\label{fig:reversal}
\end{figure}

Table~\ref{tab:instr-horizon-pairs} gives the two pairs of Section~\ref{sec:comparison}, $\piz$ and $\pif$ against SmolVLA-LP, on LIBERO-Plus Object at 50 and at 10 executed actions. Figure~\ref{fig:reversal} plots every pair of models by its margin before and after each change, and pairs in the shaded quadrants change order. Over the 45 pairs of the ten models, the repair moves 17, 17, 16 and 17 pairs by at least 10 points on Object, Spatial, Goal and Long. Each of these pairs contains SmolVLA or SmolVLA-LP, and none of the 28 pairs without them moves by that much or reverses. The repair reverses \OrdRevSpatial{} to \OrdRevObject{} pairs per suite. The execution change reverses \HEndSupAll{} of the 40 pair--suite combinations of the five models whose default is 50, and \HEndXvlaSupAll{} of the 60 once X-VLA is included. Here a reversal requires the margins at the two settings to have opposite signs with both 95\% intervals excluding zero, and the point estimates change sign in \HEndCrossAll{} and \HEndXvlaCrossAll{} of these combinations.

\begin{table}[!htbp]
\centering\small
\caption{\textbf{Margins before and after the instruction repair on LIBERO-Plus Object.} The margin is the first model's success minus that of SmolVLA-LP, over all 2{,}518 Object variants with every outcome at both horizons (not the subset of Appendix~\ref{app:pairedrng}); 95\% intervals are computed by resampling variants or base tasks (``tasks''; 10{,}000 draws each), and a pair is reversed under a resampling unit when both of its intervals exclude zero with opposite signs.}
\label{tab:instr-horizon-pairs}
\vspace{4pt}%
{\footnotesize\setlength{\tabcolsep}{2pt}\input{tables/appendix_instr_horizon_pairs}}
\end{table}

\FloatBarrier
\subsection{Resampling base tasks}
\label{app:cluster}
The 10{,}030 variants are generated from forty base tasks, so variants of one base task share its difficulty. Table~\ref{tab:ordering-counts} repeats the counts with the base task as the resampling unit: each draw takes base tasks with replacement and keeps all of their variants. On average, the intervals widen 1.8- to 3.7-fold, depending on the suite. Every pair whose margin changes by at least 10 points under the repair still has an interval that excludes zero on Object, Goal and Long, and 14 of 17 do on Spatial. Five of the seven reversals remain.

\begin{table}[!htbp]
\centering\small
\caption{\textbf{Pairs of models whose change in margin has an interval excluding zero, and reversals.} Intervals are computed by resampling variants or base tasks.}
\label{tab:ordering-counts}
\label{tab:cluster}
\vspace{4pt}%
\input{tables/app_resampling}
\end{table}

\subsection{State comparisons at the default horizon}
\label{app:inversions}
Section~\ref{sec:comparison} reports that the initial-state gaps reverse the LIBERO-Object average ordering of one pair of models. Table~\ref{tab:suite-inversion} gives the difference between every pair of the five evaluated models at their default settings, averaged over the ten tasks, at the standard initial states and on the reference-selected conditions, with percentile intervals from 20{,}000 draws that resample initial states within each task for every model and condition. The asterisk uses simultaneous two-sided 95\% max-$|T|$ intervals over the ten pairs, computed separately for the two sets with each difference studentized by its bootstrap standard deviation, together with Holm-adjusted tests at the 5\% level.

\begin{table}[!htbp]
\centering\small
\caption{\textbf{Suite-average differences between the five evaluated models.} Entries are the first model minus the second, in points. An asterisk marks a pair whose differences have opposite signs with both simultaneous intervals excluding zero after correction over the ten pairs.}
\label{tab:suite-inversion}
\vspace{4pt}%
\pendingsized{\input{tables/app_suite_pairs}}
\end{table}

\subsection{Common offset}
\label{app:residuals}
\label{app:scale}
Section~\ref{sec:comparison} summarizes each change by the largest deviation that remains after subtracting the one offset that fits all models best. This residual is half the range of the models' changes and is zero when every model responds equally. Table~\ref{tab:residuals} gives it for each change and suite, with the state gap at each model's default horizon and at ten executed actions. Without the two SmolVLA models, the instruction residual falls from up to \ResInstrMax{} points to at most \ResInstrNoSmolMax{} points, and the execution residual falls from \ResHorizonAllMin{}--\ResHorizonAllMax{} to \ResHorizonNoSmolMin{}--\ResHorizonNoSmolMax{} points, so the other models still respond by different amounts. A change shared by all models would still leave a residual from sampling at the same sample sizes.

\begin{table}[!htbp]
\centering\small
\caption{\textbf{Residual of the best common offset, in points.}}
\label{tab:residuals}
\label{tab:residual-scale}
\vspace{4pt}%
\pendingsized{\input{tables/app_residuals}}
\end{table}
\FloatBarrier

%% file: tables/appendix_instr_horizon_pairs.tex
\begin{tabular}{@{}llccccc@{}}
\toprule
Pair & & \multicolumn{2}{c}{Original margin} & \multicolumn{2}{c}{Repaired margin} & Reversed \\
\cmidrule(lr){3-4}\cmidrule(lr){5-6}
 & \multicolumn{1}{c}{$h$} & variants & tasks & variants & tasks & variant / task \\
\midrule
$\pi_0$ $-$ SmolVLA-LP & 50 & +36.5 [34.4, 38.6] & [29.0, 43.4] & $-$12.4 [$-$14.3, $-$10.3] & [$-$21.2, $-$4.2] & yes / yes \\
 & 10 & +23.4 [20.9, 25.8] & [12.1, 32.1] & $-$12.6 [$-$14.4, $-$10.8] & [$-$17.3, $-$7.9] & yes / yes \\
$\pi_{0.5}$ $-$ SmolVLA-LP & 50 & +46.9 [44.8, 49.0] & [40.2, 54.0] & $-$3.7 [$-$6.1, $-$1.3] & [$-$11.2, 4.7] & yes / no \\
 & 10 & +58.9 [56.9, 60.8] & [53.8, 63.4] & +13.3 [11.4, 15.3] & [9.5, 16.2] & no / no \\
\bottomrule
\end{tabular}

%% file: tables/app_resampling.tex
\begin{tabular}{@{}lcccc@{}}
\toprule
\multirow{2}{*}{Suite} & \multicolumn{2}{c}{Change in margin, interval excludes zero} & \multicolumn{2}{c}{Reversals} \\
\cmidrule(lr){2-3}\cmidrule(lr){4-5}
 & Variants & Base tasks & Variants & Base tasks \\
\midrule
Object & 30 & 24 & 4 & 3 \\
Spatial & 23 & 15 & 0 & 0 \\
Goal & 33 & 21 & 1 & 1 \\
Long & 25 & 22 & 2 & 1 \\
\bottomrule
\end{tabular}

%% file: tables/app_suite_pairs.tex
\begin{tabular}{@{}lll@{}}
\toprule
Pair & \multicolumn{1}{c}{Standard} & \multicolumn{1}{c@{}}{Reference-selected} \\
\midrule
$\pi_{0.5}$ $-$ SmolVLA & $+$10.8 {\footnotesize [$+$4.0, $+$17.7]} & $+$26.8 {\footnotesize [$+$22.1, $+$31.5]} \\
$\pi_{0.5}$ $-$ SmolVLA-LP & $-$1.5 {\footnotesize [$-$7.0, $+$4.0]} & $+$23.8 {\footnotesize [$+$19.3, $+$28.2]} \\
$\pi_{0.5}$ $-$ $\pi_0$ & $+$15.2 {\footnotesize [$+$8.3, $+$22.0]} & $+$31.6 {\footnotesize [$+$27.3, $+$35.8]} \\
$\pi_{0.5}$ $-$ X-VLA$^\ast$ & $-$13.3 {\footnotesize [$-$18.0, $-$9.0]} & $+$17.9 {\footnotesize [$+$13.4, $+$22.4]} \\
SmolVLA $-$ SmolVLA-LP & $-$12.3 {\footnotesize [$-$18.8, $-$5.7]} & $-$3.0 {\footnotesize [$-$7.2, $+$1.2]} \\
SmolVLA $-$ $\pi_0$ & $+$4.3 {\footnotesize [$-$3.3, $+$12.0]} & $+$4.7 {\footnotesize [$+$1.2, $+$8.4]} \\
SmolVLA $-$ X-VLA & $-$24.2 {\footnotesize [$-$30.0, $-$18.5]} & $-$8.9 {\footnotesize [$-$13.2, $-$4.7]} \\
SmolVLA-LP $-$ $\pi_0$ & $+$16.7 {\footnotesize [$+$10.3, $+$23.2]} & $+$7.8 {\footnotesize [$+$4.5, $+$11.1]} \\
SmolVLA-LP $-$ X-VLA & $-$11.8 {\footnotesize [$-$15.8, $-$8.0]} & $-$5.9 {\footnotesize [$-$10.1, $-$1.7]} \\
$\pi_0$ $-$ X-VLA & $-$28.5 {\footnotesize [$-$34.2, $-$23.2]} & $-$13.7 {\footnotesize [$-$17.3, $-$10.0]} \\
\bottomrule
\end{tabular}

%% file: tables/app_residuals.tex
\begin{tabular}{@{}lcrcrr@{}}
\toprule
Change & Models & \multicolumn{1}{c}{Object} & Spatial & \multicolumn{1}{c}{Goal} & \multicolumn{1}{c@{}}{Long} \\
\midrule
Initial state (default actions) & 5 & 15.6 & -- & -- & -- \\
Initial state (ten actions) & 5 & 33.5 & -- & -- & -- \\
Instruction & 10 & 25.7 & 14.6 & 24.3 & 19.9 \\
Instruction, without the SmolVLA pair & 8 & 1.6 & 1.2 & 4.1 & 2.1 \\
Executed actions, LIBERO & 6 & 12.4 & 7.2 & 7.9 & 11.9 \\
\bottomrule
\end{tabular}

%% file: appendix/f_toy.tex
\section{A Toy Non-identifiability Construction}
\label{app:theory}
The state comparison has a simple limiting case in which even perfect measurement on the evaluated initial states does not necessarily determine performance outside them.

\begin{proposition}
Let an evaluation observe only trajectories initialized in a set $C$. There exist deterministic dynamics and two policies whose trajectory distributions agree for all initial states in $C$, but whose success probabilities differ on states outside $C$ that another policy solves.
\end{proposition}
\begin{proof}
Consider $x_{t+1}=1.1x_t+a_t$ with absorbing failure at $|x|\geq1$, let an episode succeed when the state does not reach the failure set within a horizon of $T\geq8$ steps, and let $C=(-0.5,0.5)$. Define
\[
\pi_g(x)=-0.6x,\qquad
\pi_b(x)=
\begin{cases}
-0.6x,&|x|<0.5,\\
0,&|x|\geq0.5.
\end{cases}
\]
For every initial state in $C$, both policies induce $x_{t+1}=0.5x_t$ and remain in $C$. Their actions, states, and success outcomes therefore agree. The same holds for every metric computed from those trajectories. Let the wider evaluation distribution be uniform on $[-0.95,0.95]$. Policy $\pi_g$ contracts to the origin from every such state and succeeds with probability one. Under $\pi_b$, states with $|x_0|<0.5$ contract, while the remainder move outward and fail within eight steps. Its survival probability is $1/1.9\simeq0.526$. All states in the wider distribution are recoverable because $\pi_g$ solves them. Equality of the trajectory distributions from $C$ therefore does not identify success on the wider distribution.
\end{proof}